\documentclass[10pt,journal,compsoc]{IEEEtran}

\usepackage{amsmath,amssymb,amsfonts}
\usepackage{graphicx,float}
\usepackage{ragged2e}
\usepackage{textcomp}

\usepackage[ruled,vlined]{algorithm2e}
\usepackage{subcaption}
\usepackage{float}
\usepackage{rotate}
\usepackage{lscape}
\usepackage{rotating}
\usepackage{fancyvrb}
\usepackage{listings}
\usepackage{todonotes}
\usepackage[para]{footmisc}
\usepackage{booktabs}
\usepackage{xcolor,colortbl}
\usepackage{hyperref}
\usepackage{multirow}
\usepackage{cleveref}
\usepackage{rotating}
\usepackage{wrapfig}
\usepackage[noadjust]{cite}
\usepackage{paralist}
\usepackage{listings}
\usepackage{lipsum}
\usepackage{enumitem}
\usepackage{verbatim}
\usepackage{pdfpages}
\usepackage[many]{tcolorbox}
\usepackage[nodisplayskipstretch]{setspace}
\usepackage{xcolor,colortbl}
\definecolor{Gray}{gray}{0.85}
\definecolor{LightCyan}{rgb}{0.88,1,1}

\begin{document}

\title{A Biomedical Knowledge Graph for Biomarker Discovery in Cancer}

\author{Md. Rezaul Karim, Lina Molinas Comet, Oya Beyan, Dietrich Rebholz-Schuhmann, and Stefan Decker

\thanks{\hspace{-2mm} Rezaul Karim \& Stefan Decker are with RWTH Aachen University, Germany\\ 
\hspace{-4mm} Lina Comet is withFraunhofer FIT, Germany\\
\hspace{-4mm} Dietrich Rebholz-Schuhmann is with ZBMED - Information Center for Life Sciences and University of Cologne, Germany}
}

\markboth{}%
{Karim \MakeLowercase{\textit{et al.}}: Building a Biomedical Knowledge Graph for Biomarker Discovery in Cancer}


\maketitle

\begin{abstract}
\justifying
    Structured and unstructured data and facts about drugs, genes, protein, viruses, and their mechanism are spread across a huge number of scientific articles. These articles are a large-scale knowledge source and can have a huge impact on disseminating knowledge about the mechanisms of certain biological processes. A domain-specific knowledge graph~(KG) is an explicit conceptualization of a specific subject-matter domain represented w.r.t semantically interrelated entities and relations. A KG can be constructed by integrating such facts and data and be used for data integration, exploration, and federated queries. However, exploration and querying large-scale KGs is tedious for certain groups of users due to a lack of knowledge about underlying data assets or semantic technologies. Such a KG will not only allow deducing new knowledge and question answering~(QA), but also allows domain experts in exploration. Since cross-disciplinary explanations are important for accurate diagnosis, it is important to query the KG to provide interactive explanations about learned biomarkers. Inspired by these, we construct a domain-specific KG particularly for cancer-specific biomarker discovery. The KG is constructed by integrating cancer-related knowledge and facts from multiple sources. First, we construct a domain-specific ontology, which we call OncoNet Ontology~(ONO). The ONO ontology is developed to enable semantic reasoning for verification of the predictions for relations between diseases and genes. The KG is then developed and enriched by harmonizing the ONO, additional metadata schemas, ontologies, controlled vocabularies, and additional concept from external sources using a BERT-based information extraction method. BioBERT and SciBERT are finetuned with the selected articles crawled from the PubMed. We listed down some queries and some examples of QA and deducing knowledge based on the KG. 
\end{abstract}


\section{Introduction}
In systems medicine, holistic approaches for complete systems are applied to understand complex physiological and pathological processes, thereby forming the basis for the development of innovative methods for the diagnosis, therapy, and prevention strategies mitigating disease conditions1. Artificial intelligence~(AI) and machine learning~(ML) are widely used for the analysis and interpretation of multimodal data~(e.g., multi-omics data, imaging data, clinical outcomes, medication data, disease progression, lifestyle information, etc.) to achieve early detection, prediction and diagnosis of diseases, personalized interventions, and finally identifying biomarkers and targets for new therapies~\cite{ballester2021artificial,tran2021deep}. With the help of ML models for disease progression, the clinical data can be combined and fully analyzed in a multimodal data setting for new biomarker discovery and early disease detection. Yet, the adoption of data-driven approaches has been hampered in many clinical settings by the lack of scalable computational methods~(e.g., ML models) that can deal with large-scale data (being often heterogeneous, high dimensional, unstructured, and having high levels of uncertainty) to perform in a reliable and safe manner~\cite{phan2016integration}. The development of solutions for precision medicine requires the integration of multi-omics data (genomics, transcriptomics, proteomics, or metabolomics data), into clinical decision-making process~\cite{xie2018adaptively}. Furthermore, harmonization and seamless integration of diverse datasets from heterogeneous and distributed systems across different platforms is challenging~\cite{xie2018adaptively} as the quantity and diversity of biomedical data, and the spread of clinically relevant knowledge across multiple biomedical databases and publications, pose a challenge to data integration.

A challenging scenario for AI solutions is the use of large-scale clinical data for improving the diagnosis and treatment of medical conditions based on cancer~\cite{ballester2021artificial,tran2021deep}. Cancer is characterized as a heterogeneous disease, having many types and subtypes~\cite{karim2019onconetexplainer,karimACCA2019}. It is caused when cells turn abnormal, divide rapidly, and spread to other tissues and organs and may be further driven by a series of genetic mutations of genes induced by selection pressures of carcinogenesis in the cells. The so-called marker genes including oncogenes and tumor suppressor genes are often responsible for cancer growth. When a gene is over- or under-expressed as a differentially expressed gene, it becomes uncontrollable proliferation or immorality of cancer cells~\cite{xie2018adaptively,tran2021deep}. Although the difference in the average of expression values between two sample classes is frequently employed in transcriptomics analyses, such difference is not the only way a gene can be expressed differentially~\cite{xie2018adaptively}. With more than 200 different types identified to date, cancer has become the second leading cause of death worldwide [10]. According to the National Cancer Institute2 an estimated 17.35 million new cancer cases were diagnosed in the United States in 2018 of which 609,640 people died; while there were 18 million new cases of which 9.5 million deaths were reported worldwide. The number of new cases is expected to rise to 23.6 million by 2030 and is anticipated to increase by 70\% by 2035 and to 29.5 million by 2040. On the other hand, according to Federal Statistical Office3, 231271 people died of cancer in Germany in 2020, where breast and lung cancer are the leading causes of death\footnote{\url{https://www.cancer.org/research/cancer-facts-statistics/all-cancer-facts-figures/cancer-facts-figures-2018.html}}.

Cancer research relies on a large number of diverse datasets generated by different omics technologies. Electronic health record data, which are gathered at the point-of-care, are increasingly utilized in pre-clinical and clinical research as well as health management and analyzed in unison with genomic data. Curation, management, and analysis of these data present unique challenges arising from data heterogeneity, complexity, and size. Ontologies and knowledge graphs are being increasingly adopted to address these issues, and have great potential to support multidisciplinary cancer research efforts. Moreover, ontology-based approaches for genomics, machine learning, and clinical decision support are a growing hot topic due to their potential to support explainability.

Experts are often interested in gathering and comprehending knowledge and mechanism of certain biological process, e.g., diseases to design strategies in order to develop prevention and therapeutics decision making process. \textit{``Knowledge is something that is known and can be written down''}~\cite{nonakatakeuchi1995}. Knowledge containing simple statements, e.g., \textit{``TP53 is an oncogene"} or quantified statements, such as \textit{``All oncogenes are responsible for cancer"} can be extracted from structured sources such as knowledge or rule bases. Moreover, knowledge can be extracted from external sources like scientific articles, where KG could be an effective means to capture facts from heterogeneous data sources. For example, scientific literature and patents provide a huge treasure of structured and unstructured information about different biological entities. One prominent example is PubMed, which contain millions of scientific articles is a great source of knowledge in biomedical domain~\cite{xu2020building}. PubMed data are mostly unstructured and heterogeneous. This makes the knowledge extraction process very challenging. 

Domain experts may need to rely on up-to-date findings from external sources, e.g., knowledge and facts about drugs, genes, protein, and their mechanism are spread across a huge number of structured~(knowledge bases) and unstructured~(e.g., scientific articles)~\cite{karim_phd_thesis_2022,karim2022question}. 
In the last few years, thanks to the availability of sizeable open-access article repositories such as PubMed Central (Beck 2010), arxiv (https://arxiv.org) bioarxiv (https://www.biorxiv.org/) as well as ontology databases which hold entities and their relations (Lambrix et al. 2007), the research community has focused on text mining tools and machine learning algorithms to digest these corpora and extract valuable semantic knowledge from them. These articles form a large-scale biomedical data source and can have a huge impact on disseminating mechanisms about certain biological processes, e.g., diseases to design strategies o develop prevention and therapeutics decision-making process~\cite{karim2022question}. 

However, extracting knowledge and facts from unstructured, heterogeneous, and scattered sources is a very challenging task. The problem of semantic heterogeneity is further compounded due to the flexibility of semi-structured data and various tagging methods applied to documents or unstructured data. 
Semantic Web~(SW) technologies address data variety, by proposing graphs as a unifying data model, to which a data can be mapped in the form of a graph structure. Ontology-based named entity extraction and disambiguation help with unambiguous identification of entities in heterogeneous data and assertion of applicable named relationships that connect these entities together. A graph may not only contain data, but also metadata and domain knowledge~(ontologies containing axioms or rules), all in the same uniform structure, and are then called knowledge graph~(KGs)~\cite{wilcke2017knowledge,hogan2020knowledge}. Owing to the SW technologies that offer functionality to connect previously isolated pieces of data and knowledge, associate meaning to them, and represent knowledge extracted from them. In particular, ontology-based named entity relation extraction~(NER) and disambiguation help with unambiguous identification of entities in heterogeneous data and assertion of applicable named relationships that connect these entities together. 

Information extraction is the process of automatically extracting structured knowledge and facts from such unstructured and/or semi-structured machine-readable documents and other electronically represented sources~\cite{Liddy.2001}. Information extraction is typically done in three steps: i) \emph{NER} that recognizes the named entities scientific articles, ii) \emph{entity linking}~(EL) to link extracted named entities, and iii) \emph{relation extraction}~(RE). NER is recognizing domain-specific proper nouns in a biomedical corpus. It can be performed by fine-tuning a domain-specific BERT variant such as BioBERT~\cite{BioBERT} and SciBERT~\cite{SciBERT} on relevant articles. The RE step also involves relation classification, which is typically formulated as a classification problem to classify the relationship between  entities identified in the text~\cite{xue2019fine}. A classifier takes a piece of text and two entities as inputs and predicts possible relations between the entities as output. The EL task is about associating mentions of the named entities in a text with the existing nodes of our target KG~\cite{hogan2020knowledge}. For example, for the abstract: \emph{``Cyclooxygenase~(COX)-2 mRNA and protein expression were found to be frequently elevated in human pancreatic adenocarcinomas and cell lines derived from such tumours. Immunohistochemistry demonstrated cytoplasmic COX-2 expression in 14 of 21~(67\%) pancreatic carcinomas. The level of COX-2 mRNA was found to be elevated in carcinomas, relative to the histologically normal pancreas from a healthy individual, as assessed by reverse transcription-PCR.''}, a NER model would be able to recognize named entities classified as diseases, chemical or genetic, as shown in, where named entities are highlighted with different HMs.  

Facts containing simple statements, e.g., \textit{``TP53 is an oncogene"} or quantified statements:\textit{``oncogenes are responsible for cancer"} can be extracted and integrated into a knowledge graph~(KG). A simple statement can be accumulated as an edge in a KG, while quantified statements provide a more expressive way to represent knowledge, which however requires \textit{ontologies}~\cite{hogan2020knowledge}. Hogan et al.~\cite{hogan2020knowledge} defined KG as \textit{``a graph of data intended to accumulate and convey knowledge of the real world, whose nodes represent entities of interest and whose edges represent potentially different relations between these entities''}. Nodes in a KG represent entities and edges represent binary relations between those entities~\cite{hogan2020knowledge}. A KG can be defined as $G=\{E,R,T\}$, where $G$ is a labelled and directed multi-graph, and $E, R, T$ are the sets of entities, relations, and triples, respectively and a triple can be represented as $(u,e,v) \in T$, where $u \in E$ is the head node, $v \in E$ is the tail node, and $e \in R$ is the edge connecting $u$ and $v$~\cite{hogan2020knowledge}.

\begin{figure*}[h]
	\centering
	\includegraphics[width=0.8\textwidth]{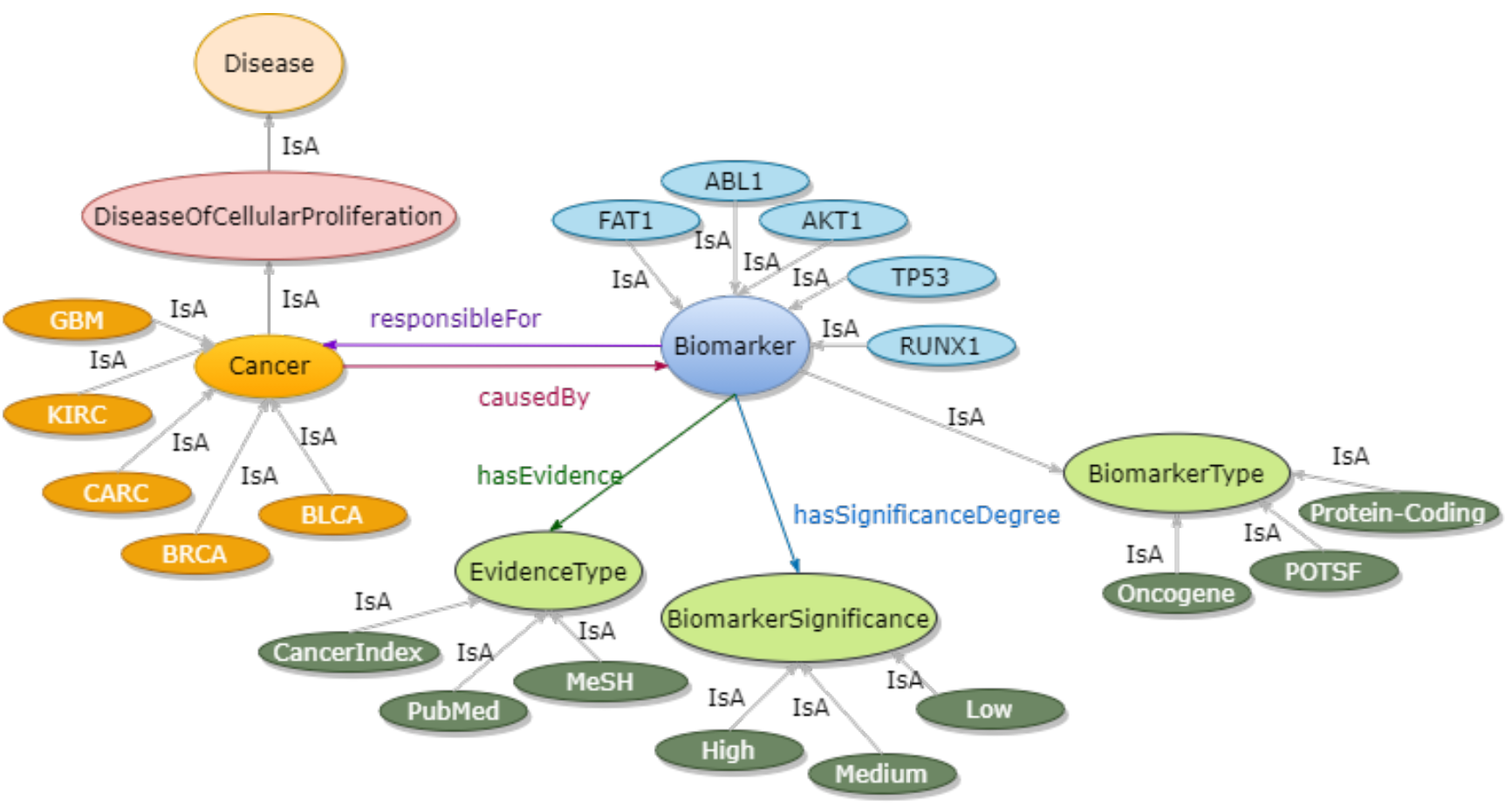}
	\caption{An example knowledge graph that connects knowledge, facts, and data about cancer~\cite{karim_phd_thesis_2022}} 
	\label{fig:cancer_ontology_example}
\end{figure*}

A common way to represent the extracted facts in Resource Description Framework~(RDF). RDF is a semantic data model, where the linking structure of a KG forms a directed graph and triples are represented in the form of $(u,e,v)$ or $(subject,predicate,object)$. Each triple forms a connected component of a sentence for a KG. A number of languages have been proposed for querying a KG in RDF format~\cite{hogan2020knowledge}, including the SPARQL query language for RDF graphs\footnote{\url{https://www.w3.org/TR/sparql11-query/}}.
Reasoning over KGs enables consistency checking to recognize conflicting facts, classification by defining taxonomies, and deductive inferencing by revealing implicit knowledge from a set of facts~\cite{futia2020integration}. However, exploration, processing, and analysis of large-scale KGs pose a great challenge to current computational methods. To provide cancer diagnosis reasoning over the DNN models, an integrated domain-specific KG is required, which is subject to the availability of an efficient NLP-based information extraction method and a domain-specific ontology. 

\section{Related work}\label{chapter_8:rw}
Research initiatives are gradually adopting SW technologies~\cite{karim2018improving} such as knowledge bases~(KBs) and domain-specific ontologies as the means of building structured networks of interconnected knowledge~\cite{futia2020integration}. Hogan et al.~\cite{hogan2020knowledge} have provided a comprehensive review of articles on KGs, covering knowledge graph creation, enrichment, quality assessment, and refinement. Apart from these literature that focuses on the theoretical concepts, several large-scale KGs have been constructed either by manual annotation, crowd-sourcing~(e.g., DBpedia) or by automatic extraction from unstructured data~(e.g., YAGO)~\cite{wang2015explicit} targeting KG analytics for specific use cases. 
Life sciences is an early adaptor of SW technologies. Numerous research efforts from the scientific communities have focused on constructing large-scale KGs for life science research. For example, Bio2RDF and PubMed KG~\cite{xu2020building} are developed to accelerate bioinformatics research. The former integrates 35 life sciences datasets such as dbSNP, GenAge, GenDR, LSR, OrphaNet, PubMed, SIDER, WormBase, contributing 11 billion RDF triples. 

Alshahrani et al.~\cite{alshahrani2017neuro}, built a biological KG based on the gene ontology~(GO), human phenotype ontology~(HPO), and disease ontology. Then they performed feature learning over the KG. Their method combines knowledge representation using symbolic logic and automated reasoning, with neural networks to generate embeddings of nodes. Then the learned embeddings are used in downstream application development such as link prediction, finding candidate genes of diseases, protein-protein interactions, and drug target relation prediction. Hasan et al.~\cite{hasan2020knowledge} developed a prototype KG based on the Louisiana Tumor Registry dataset\footnote{https://sph.lsuhsc.edu/louisiana-tumor-registry/}. Their approach provides scenario-specific querying, schema evolution for iterative analysis, and data visualization. Although the KG found to be effective at population-level treatment sequences, it is built on a limited data setting, does not provide comprehensive knowledge about cancer genomics for the majority of the cancer types.

Hu et al.~\cite{hu2015semantic} developed a breast cancer KG by integrating: i) Dutch medical guidelines\footnote{A semantic representation of Dutch medical guidelines.} that contain conclusions and their evidences with UMLS and SNOMED CT medical terminologies, ii) genomic data of female patients, iii) clinical trials from official NCT website, iv) semantic annotations w.r.t eligibility criteria generated with XMedLan\footnote{A Xerox NLP tool for medical text.}, iv) selected medical publications from PubMed released by the Linked Life Data\footnote{A semantic data integration platform for biomedical domain.}. 
Although these KGs can be used to explore the relationship among knowledge and data sources of breast cancer and support for clinical DSS, no comprehensive KBs or KGs have been developed targeting a wide variety of cancer types to date. 

In biomedical treatment protocols, structured data is first extracted from various unstructured sources such as pathology reports, radiology reports, and medical records, before storing them in a database for reporting and diagnosis purposes. In hospital information systems or electronic health records, the cancer registry data is typically categorized for each reported cancer case or tumor at the time of diagnosis~\cite{hasan2020knowledge}. Besides, it may contain demographic information about the patient such as age, gender, location at time of diagnosis, planned and completed primary treatment information, and survival outcomes~\cite{hasan2020knowledge}. Cancer-related scientific articles that report unstructured data also provide facts and knowledge about cancer characteristics, treatments, and outcomes. Extracting information from structured data is straightforward, extracting the same from unstructured data maybe highly domain specific. 

A concrete example is cancer-specific biomarker. Shen et al.~\cite{POSTF} have reported that some genes have both oncogenic and tumor-suppressor functionality called proto-oncogenes with tumor-suppressor function~(POTSF). The majority of POTSF genes act as transcription factors or kinases and exhibit dual biological functions, e.g., they both positively and negatively regulate transcription in cancer cells~\cite{POSTF}. Besides, specific cancer types like leukemia are over-represented by POTSF genes, whereas some common cancer types like lung cancer are under-represented by them~\cite{POSTF}. Apart from the oncogenes and protein-coding genes, proto-oncogenes are a group of genes that cause normal cells to become cancerous when they are mutated~\cite{slamon1987proto}. Oncogenes result from the activation of proto-oncogenes, whereas a POSTF itself cause cancer when they are inactivated. For example, TP53 is a POTSF and abnormalities of TP53 have been found in more than half of human cancers\footnote{\url{www.cancer.org/cancer/cancer-causes/genetics/genes-and-cancer/oncogenes-tumor-suppressor-genes.html}}. Mutations in proto-oncogenes are typically dominant and the mutated version of a proto-oncogene is called an oncogene~\cite{slamon1987proto}. 

Since structured and unstructured data about cancer is constantly expanding and evolving, linking large-scale heterogeneous data sources is challenging, as heterogeneous data sources often do not follow any common data-storing standard~\cite{hasan2020knowledge}. Feature- or kernel-based methods such as the conditional random field~(CRF) are earlier approaches used for the NER and RE in a pipeline setting~\cite{xue2019fine}. However, performance of traditional approaches heavily depends on manual feature engineering, domain-specific knowledge, and a deep understanding of linguistics~\cite{xue2019fine}. Therefore, approaches based on DNNs have emerged that are not only capable of automatic extraction of structured knowledge, but also leverage end-to-end learning capability~\cite{dogan2019fine}. 

Lately, several deep architectures based on recurrent types of neural networks have been proposed for entity recognition and RE. Approaches based on Bidirectional Long Short Term Memory~(Bi-LSTM) or Gated Recurrent Unit cells are more widely used~\cite{zheng2017joint}. For example, Geng et al.~\cite{zhang2020semi} used a Bi-LSTM network with a subsequent CRF, where inputs to the Bi-LSTM are concatenated word representations that combine pre-trained contextual word embeddings using embeddings from language models~(ELMo)~\cite{peters2018deep}, character-level word representations, and positional features of the word in the document. Joint entity and REn approaches are also proposed. Zhang et al.~\cite{zheng2017joint} used joint-BiLSTM -a joint tagging method that can transform both tasks into sequence tagging through parameter sharing in an end-to-end learning setting. Although Bi-LSTM approaches outperform CRF-based approaches, their shortcoming lies in the incapability of semantic annotation and validation of the extracted entities and relations, i.e., not all relations can be represented by triggered words~\cite{sun2020biomedical}. 

Recently, transformer language models have become the de-facto standard for representation learning in NLP, as it is feasible to perform domain-specific fine-tuning in a transfer learning setting. Compared to transformer models that benefit from abundant knowledge from pre-training and strong feature extraction capability, approaches based on Bi-LSTM have a lower generalization performance~\cite{xue2019fine}. Bidirectional Encoder Representations from Transformers~(BERT)~\cite{devlin2018bert} is a language model that utilizes the bidirectional attention mechanism and large-scale unsupervised corpora to obtain effective context-sensitive representations of words in a sentence~\cite{xue2019fine}. Several variants of BERT are proposed and widely used as transformer language models for a variety of downstream NLP tasks like NER. 

Sun et al.~\cite{sun2020biomedical} employed a transformer for predicting entity labels, where word representations are obtained from a pre-trained BERT model. BERT-based approaches have also been applied to scientific texts~(e.g, SciBERT~\cite{Beltagy2019SciBERT}) and adapted in other domains such as ChemBERTa~\cite{chithrananda2020chemberta} in chemical engineering for molecular property prediction.
Xu et al.~\cite{xu2020building} have shown that bio-entity extraction based on BioBERT~\cite{lee2020biobert} can significantly outperform DNN-based methods. This makes transformer-based language models of a great potential for tasks involving information extraction, information retrieval. Subsequently, SW technologies can be more effective for extracting knowledge from unstructured texts with high accuracy~\cite{anantharangachar2013ontology}. 
In other words, an ontology-guided information extraction method needs to be combined with state-of-the-art transformer language models. 

\section{Methods}\label{chapter_8:mm}
We construct the knowledge graph~(KG) by integrating cancer-related knowledge and facts from multiple sources. First, we construct a domain-specific ontology, which we call OncoNet Ontology~(ONO). The ONO ontology is developed to enable semantic reasoning for verification of the predictions for relations between diseases and genes. The KG is then developed and enriched by harmonizing the ONO, additional metadata schemas, ontologies, vocabularies, and additional concept properties from external sources. 

\begin{figure*}[h]
	\centering
	\includegraphics[width=0.6\textwidth]{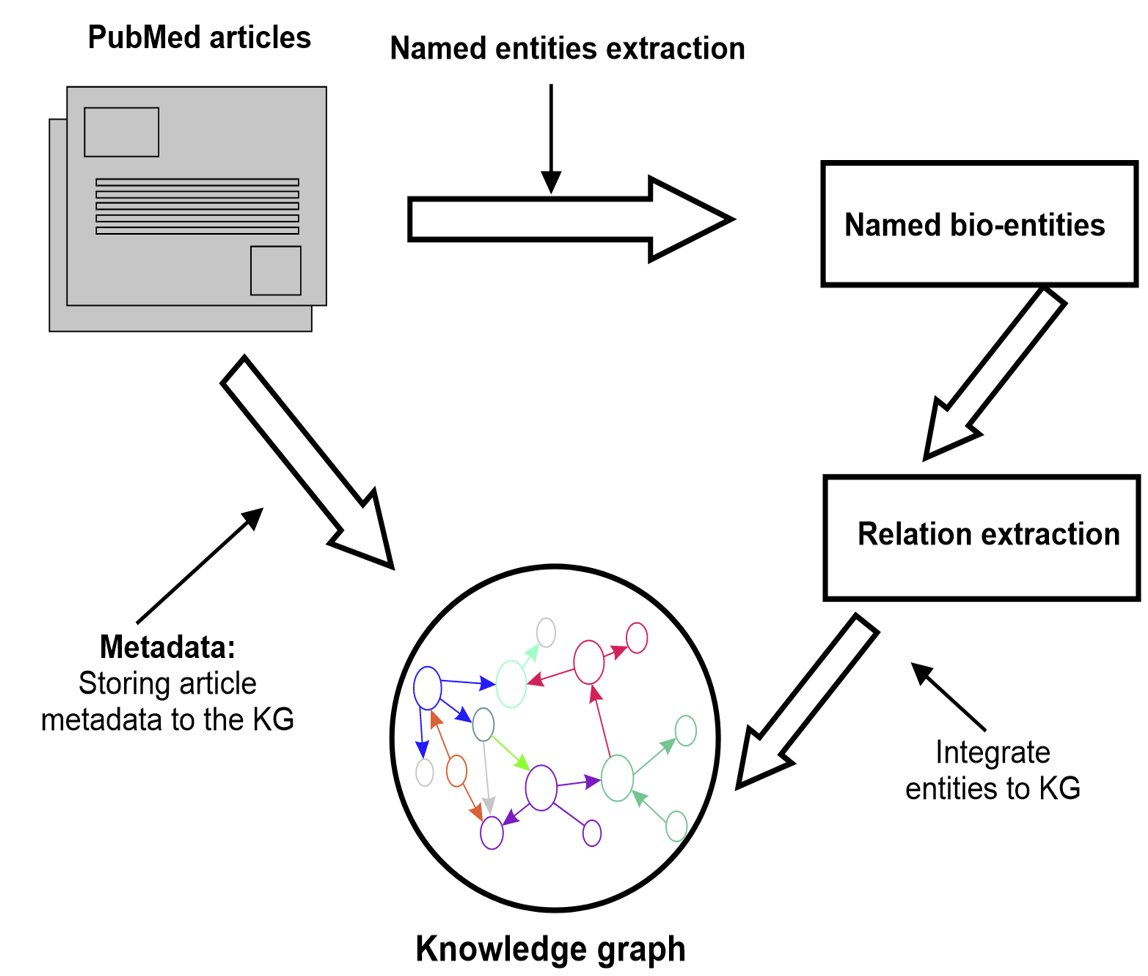}
	\caption{Workflow of our overall overall knowledge graph construction process~\cite{karim_phd_thesis_2022}} 
	\label{fig:kg_creation}
\end{figure*} 

\subsection{Ontology modelling}
Instead of building from scratch, the ONO ontology is developed based on existing ontologies. We reuse existing ontologies to represent domain knowledge about diseases, annotations about genes, and metadata~(e.g, labels, provenance). As controlled vocabularies\footnote{Concepts and their relations w.r.t cancer.} and  
facts\footnote{Factual knowledge about cancer subjects from literature.} are required, 
we considered Cancer, Biomarker, and Feature classes as the basis for the ontology: 
The \texttt{ono:Cancer} is a conceptual umbrella term that contains 33 different types of cancer types as shown in \cref{table:cancer_types}, where entities are either real biological or cancer-specific domain terms. 

\begin{figure*}[h]
	\centering
	\includegraphics[width=0.6\textwidth]{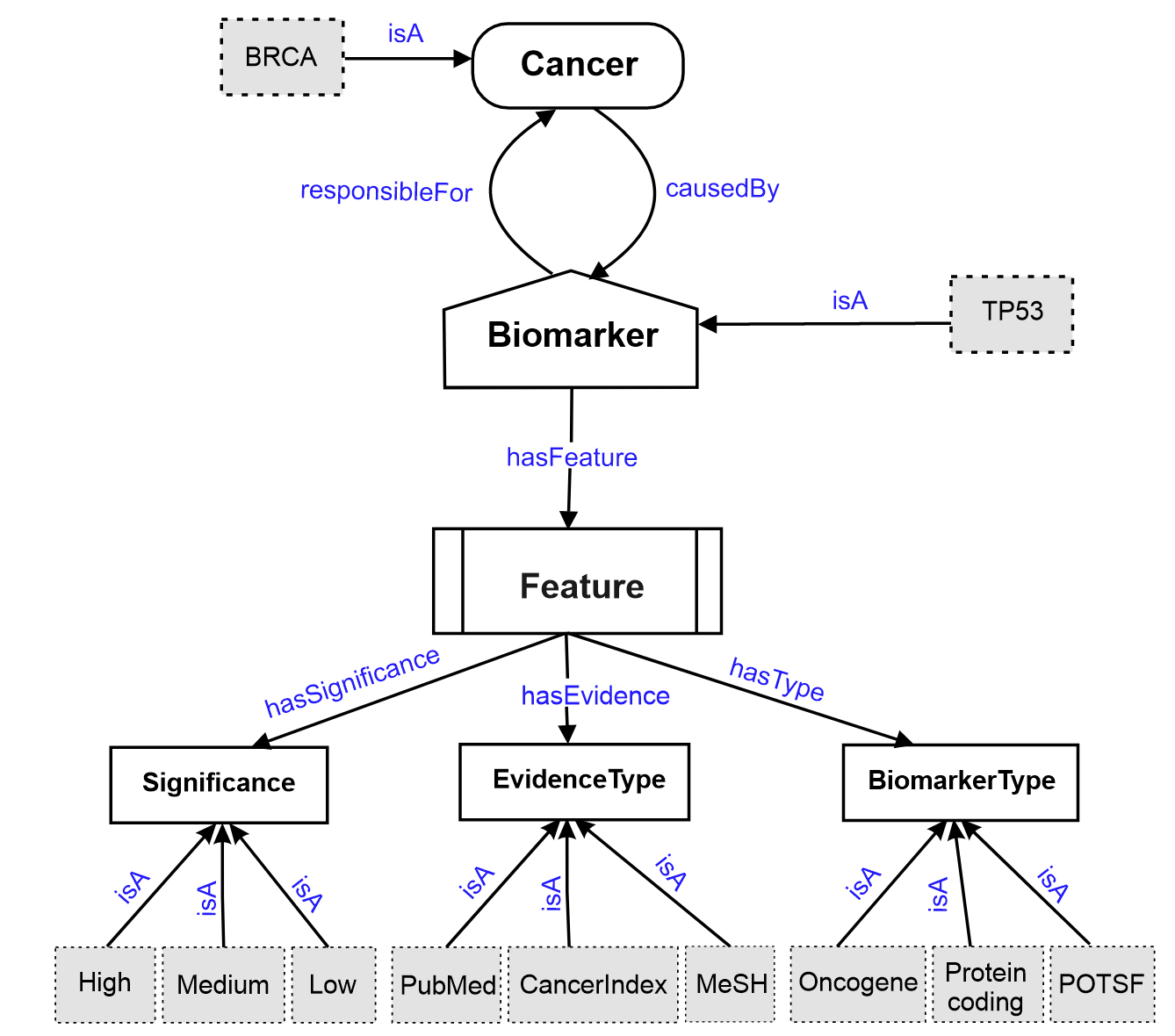}
	\caption[Cancer, Biomarker, and Feature classes and their properties that relate them]{Cancer, Biomarker, and Feature classes and their properties that relate them. The responsibility properties are shown as predicate. The dashed gray boxes signify example properties of the entities under each class~\cite{karim_phd_thesis_2022}} 
	\label{fig:3classes_full}
\end{figure*}
 
\begin{table}
\centering
    \caption[Different cancer types for the class Cancer]{Different cancer types for the class Cancer. The acronyms are used in the ontology, while the full name of the carcinoma types are used for the rule interpretations~\cite{karim2019onconetexplainer}}
    \label{table:cancer_types}
    \vspace{-2mm}
    \begin{tabular}{l|l}
        \hline
        \rowcolor{Gray}
         \textbf{Cohort} & \textbf{Carcinoma type} \\\hline
            ACC  & Adrenocortical carcinoma \\\hline
            BLCA & Bladder urothelial carcinoma \\\hline%
            BRCA & Breast invasive carcinoma \\\hline
            CESC & Cervical and endocervical cancers	\\\hline%
            CHOL & Cholangio carcinoma \\\hline 
            COAD & Colon adenocarcinoma \\\hline 
            DLBC & Diffuse large-Bcell lymphom  \\\hline
            ESCA & Esophageal carcinom \\\hline
            GBM  & Glioblastoma multiforme \\\hline
            HNSC & Head and Neck squamous cell carcinoma \\\hline
            KICH & Kidney Chromophobe  \\\hline
            KIRC & Kidney renal clear cell carcinoma \\\hline
            KIRP & Kidney renal papillary cell carcinoma	\\\hline%
            LAML & Acute Myeloid Leukemia \\\hline 
            LGG  & Brain Lower Grade Glioma \\\hline 
            LIHC & Liver hepatocellular carcinoma  \\\hline
            LUAD & Lung adenocarcinoma \\\hline
            LUSC & Lung squamous cell carcinoma \\\hline
            MESO & Mesothelioma \\\hline
            OV   & Ovarian serous cystadenocarcinoma  \\\hline
            PAAD & Pancreatic adenocarcinoma \\\hline
            PCPG & Pheochromocytoma and Paraganglioma \\\hline
            PRAD & Prostate adenocarcinoma  \\\hline
            READ & Rectum adenocarcinoma \\\hline
            SARC & Sarcoma	\\\hline%
            SKCM & Skin Cutaneous Melanoma \\\hline 
            STAD & Stomach adenocarcinoma \\\hline 
            TGCT & Testicular Germ Cell Tumors  \\\hline
            THCA & Thyroid carcinoma \\\hline
            THYM & Thymoma \\\hline
            UCEC & Uterine Corpus Endometrial Carcinoma \\\hline
            UCS  & Uterine Carcinosarcoma  \\\hline
            UVM  & Uveal Melanoma \\\hline
    \end{tabular}
\end{table}

The \texttt{ono:Biomarker} is a conceptual umbrella term that contains 660 genes that are responsible for different types of cancer~(defined in \texttt{ono:Cancer} class). A gene maybe responsible for a single or multiple types of cancer, e.g., TP53 is responsible for breast~(BRCA), ovarian~(OV), medulloblastoma~(MED), and prostate~(PRAD) cancer. For \emph{geneType}, `Oncogene', `Protein-coding', `POTFS' are possible types; for \emph{evidenceType}, `PubMed', `MeSH', `CancerIndex' are possible sources; for \emph{hasSignificance} a gene can have one of `HIGH', `MEDIUM', `LOW' levels of significance; \emph{crossResponsibility} can have one or more than one of `ACC', `BLCA', `BRCA', `CESC', `CHOL', `COAD', `DLBC', `ESCA', `GBM', `HNSC', `KICH', `KIRC', `KIRP', `LAML', `LGG', `LIHC', `LUAD', `LUSC', `MESO', `OV', `PAAD', `PCPG', `PRAD', `READ', `SARC', `SKCM', `STAD', `TGCT', `THCA', `THYM', `UCEC', `UCS', `UVM' 33-cancer types; for \emph{hasCitations}, there is at least 1 article indexed in `PubMed', `MeSH', `CancerIndex'. 

The \texttt{ono:Feature} characterize additional information about the genes~(defined in \texttt{ono:Biomarker} class) such as biomarker types, degree of significance w.r.t certain cancer types, and source of evidence. A biomarker could be one of three different types such as oncogenes, protein-coding, and POTSF. Each gene can have one of three properties as high, medium, and low according to it's level of significance. For example, TP53 is highly significantly mutated in the breast~(BRCA) and ovarian~(OV) cancer types, significantly mutated in the prostate~(PRAD) cancer type, and near significantly mutated in medulloblastoma~(MED) cancer types. These findings are evidently found in scientific articles cited in PubMed, CancerIndex, and MeSH. 

\begin{figure*}
	\centering
	\includegraphics[width=\textwidth]{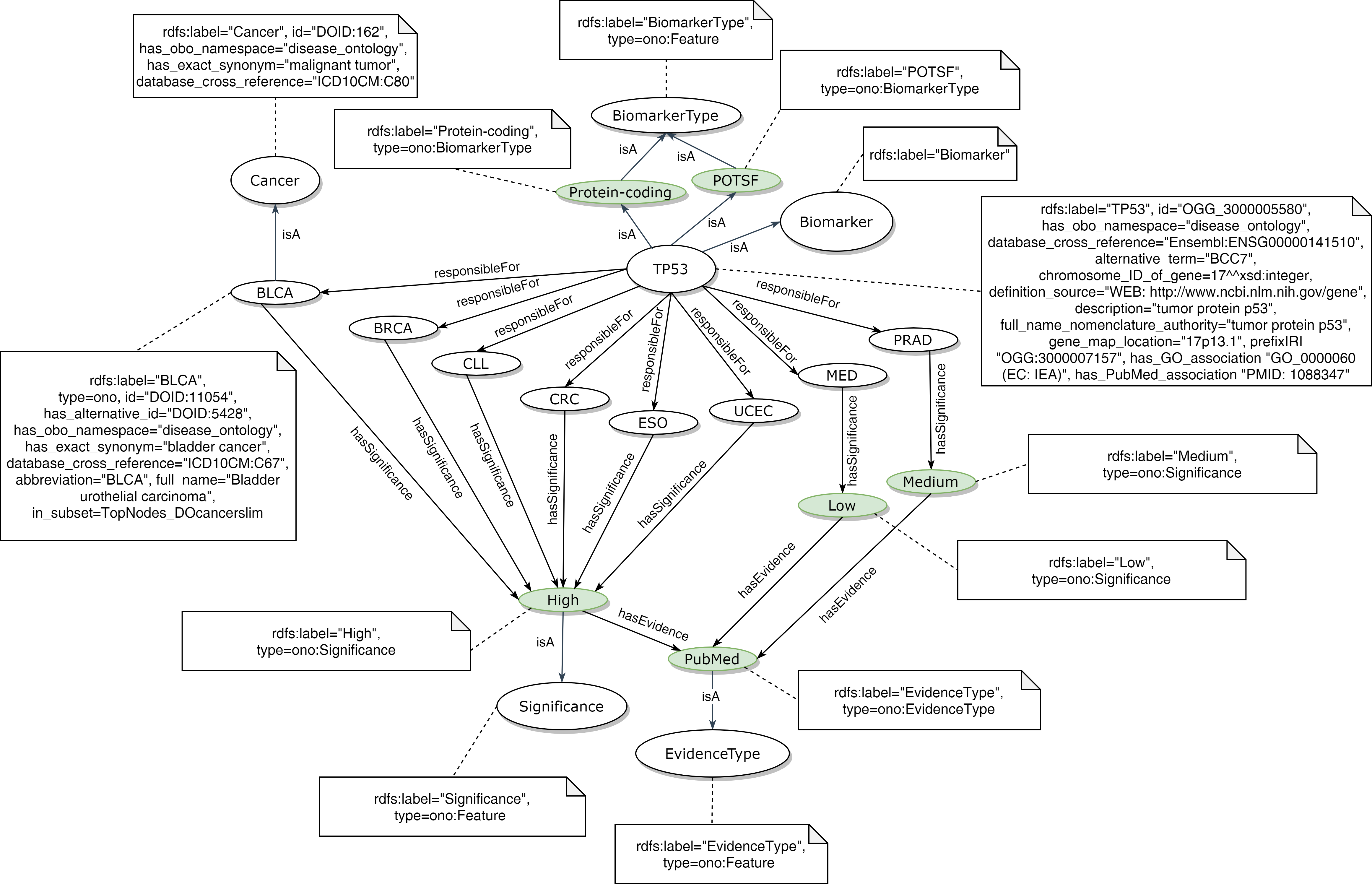}
	\caption[Properties and relevance of biomarkers w.r.t different types of cancer]{Properties of TP53 biomarker show how it is related to different types of cancer. Dashed lines indicate literals and annotations from DOID, OGG, and external sources~\cite{karim2022explainable}}
	\label{fig:tp53_v2}
	\vspace{-2mm}
\end{figure*}

We used the ontology of genes and genomes~(OGG)\footnote{\url{https://bioportal.bioontology.org/ontologies/OGG}} and the human disease ontology~(DOID)\footnote{\url{https://bioportal.bioontology.org/ontologies/DOID}} as the basis of our ontology. These two ontologies model entities about genes and genomes, and human diseases, respectively. Subsequently, We inherited the metadata of the entities from both ontologies. \Cref{fig:3classes_full} shows how the classes relate to each other using different properties, while \cref{fig:tp53_v2} shows how a biomarker, for example, TP53 is conceptualized by interconnecting three different classes.

As shown in \cref{fig:kg_creation}, we inherited the semantic concepts~(e.g., annotations, metadata of the entities, etc.) from several sources to model our ontology, which can reflect the structural aspects of the connection between biomarkers~(e.g., oncogenes, POTSF, ProteinCoding) and cancer types~(e.g. breast, colorectal cancer). In particular, gene ontology~(GO), human phenotype ontology~(HPO), disease ontology~(DO), breast cancer ontology~(BCO), cancer genetics web, TumorPortal\footnote{\url{www.tumorportal.org}}, and cancer genetic website\footnote{\url{www.cancer-genetics.org/}}.

We map all the protein identifiers to Entrez gene and are used to represent genes, proteins, and other biological entities from the DO. DO is a standardized ontology for human disease, which aims to provide consistent, reusable and sustainable descriptions of human disease terms, phenotype characteristics, and medical vocabulary disease concepts. DO semantically integrates disease and medical vocabularies via cross mapping of DO terms from MeSH, ICD, NCI, SNOMED, and OMIM. DO also provides an integrated information from several data source and analysis of the literature using data from PubMed and CancerIndex.org. Further, it also provides and integrated view of the latest research abstracts relating to genes, proteins and cancer.   

The BCO from European Bioinformatics Institute~(EBI)\footnote{\scriptsize{\url{https://www.ebi.ac.uk/ols/ontologies/doid/terms?obo_id=DOID:1612}}} provides domain knowledge about breast cancer and a rich source of carcinogenics containing relations between biological entities. Altogether, we treat biological entities like types of genes, proteins, and diseases as instances. Classes from the DO are also treated as instances. The general structure of the ontology, depicting the hierarchical relations between different classes and subclasses is shown in \cref{fig:tp53_v2}. 

There are two distinct types of entities: biological entities and classes from biomedical ontologies that provide background knowledge about entities. Each instance $f_i$ in the KG is assigned a unique IRI. Ontology-based annotations are expressed by asserting a relation between instances, e.g., a gene, cancer type, and an instance of a class. For example, we express that gene $AKT1$ has the GO association \texttt{GO:0000060} by two axioms \texttt{hasGOAssociation(AKT1,$f_1$)} and \texttt{instanceOf($f_1$, GO:0000060)}: $AKT1$ and $f_1$ are instances, \texttt{GO:0000060} is a class \url{http://purl.obolibrary.org/obo/GO\_0000060} in GO, \texttt{hasGOAssociation} is an object property, and \texttt{instanceOf} is the \texttt{rdf:type} property in OWL standard. 



\subsection{Knowledge graph construction and enrichment}
A domain-specific KG is an explicit conceptualization of a specific subject-matter domain represented w.r.t semantically interrelated entities and relations~\cite{abu2020domain}. Since cross-disciplinary explanations are important for accurate diagnosis, we aim to query the KG to provide interactive explanations about learned biomarkers. Depending on the hierarchy of abstraction, relations between concepts can be more complicated, making it difficult to interpret. 
The ONO ontology is further enriched with domain-specific knowledge and facts about cancer from  scientific literature. Annotations help enrich the KG with domain knowledge about relations between concepts to facilitate more complex and elaborate reasoning. Thus, additional facts and knowledge from scientific articles are used to enrich the KG. For example, Shen et al.~\cite{POSTF}, through database search and literature annotation, have identified 83 POTSF genes: BRCA1, CAMTA1, CBFA2T3, CDX2, CREB3L1, CREBBP, DDB2, DNMT1, DNMT3A, ETV6, EZH2, FOXA1, FOXL2, FOXO1, FOXO3, FOXO4, FOXP1, FUS, IRF4, KLF4, KLF5, NCOA4, NOTCH1, NOTCH2, NOTCH3, NPM1, NR4A3, PAX5, PML, PPARG, RB1, RUNX1, SMAD4, STAT3, TCF3, TCF7L2, TP53, TP63, TRIM24, WT, ZBTB16, BCR, CHEK2, EPHA1, EPHA3, EPHB4, FLT3, MAP2K4, MAP3K4, MST1R, NTRK3, PRKAR1A, PRKCB, SYK, ARHGEF12, BCL10, BRCA2, CBL, CDC73, CDH11, CDKN1B, DCC, DDX3×, DICER1, FAS, FAT1, GPC3, IDH1, IKZF2, LIFR, NF2, NUP98, PHF6, PTPN1, PTPN11, RHOA, RHOB, SH2B3, SLC9A3R1, SOCS1, SPOP, SUZ12, WHSC1L1. 

To extract such valuable information, a BERT-based information extraction method is employed. BioBERT and SciBERT variants of BERT are finetuned with the selected articles crawled from the PubMed, IEEE Digital Library, and Google Scholar. Our hypothesis is that, in learning to recover masked tokens, the model forms a representational topology of cancer-related unstructured sources will enable the ontology-guided knowledge extraction to outperform CRF and Bi-LSTM-based approaches for NER, entity-relationship extraction, and multi-type normalization tasks. 

\subsubsection{Article collection and preprocessing}
Cancer-related scientific articles are collected to extract information about cancer specific biomarkers from, based on the inclusion and exclusion criteria in \cref{table:inclusion_exclusion_article}. For example, some genes exhibit both oncogenic and tumor-suppressor functionality~\cite{POSTF}. The KG can be enriched by extracting the name and properties of these genes. 
Since systematic reviews of complex evidence cannot rely solely on protocol-driven search strategies~\cite{karim2018improving}, our search begins with the use of search queries with search terms and a Boolean operator such as (``cancer"[All Fields]) AND (``biomarkers"[All Fields]) and combining it with the snowball sampling searches. PubMed, MeSH, CancerIndex, IEEE Digital Library, and Google Scholar are considered as the main sources. However, only PubMed is used as it is a more comprehensive source of biomedical literature. \Cref{table:inclusion_exclusion_article} shows the statistics of the systematic searching in which we specify more recent years~(i.e., 2020-2022). While using the protocol-based and snowball sampling-based searching, only one reason was recorded for each record, albeit multiple reasons were applicable. Then we included only those articles that are marked as ``Free PMC Article". 

\begin{table*}[h!]
    \centering
    \caption{Cancer related article searching queries and related statistics~\cite{karim_phd_thesis_2022}}
    \label{table:inclusion_exclusion_article}
    \vspace{-2mm}
    \begin{tabular}{p{0.85cm}|p{14cm}|p{0.95cm}|p{0.8cm}} 
        \hline
        \textbf{Query} & \textbf{Query string} & \textbf{\#hits} & \textbf{\#Used} \\ 
        \hline
        Q1  & (``cancer"[All Fields] AND ``biomarkers"[All Fields]) OR (``oncology"[All Fields] AND ``biomarkers"[All Fields]) OR (``cancer genomics"[All Fields] AND ``biomarkers"[All Fields]) & 94,962 & 6,000 \\ 
        \hline
        Q2  & (``oncology"[All Fields] AND ``biomarkers"[All Fields]) OR (``cancer genomics"[All Fields] AND ``biomarkers”[All Fields]) OR(``cancer"[All Fields] AND ``oncogenes"[All Fields]) & 8,3841 & 4,550 \\ 
        \hline
        Q3  & (``oncology"[All Fields] AND ``oncogenes"[All Fields]) OR (``cancer"[All Fields] AND ``biomarkers"[All Fields]) OR (``cancer genomics"[All Fields] AND ``biomarkers"[All Fields]) & 87,635 & 2,250 \\
        \hline
        Q4  & (``oncology"[All Fields] AND ``biomarkers"[All Fields]) OR (``cancer"[All Fields] AND ``cancer genomics"[All Fields]) OR (``cancer"[All Fields] AND ``biomarkers"[All Fields]) & 79,960 & 3,290 \\ 
        \hline
        Total  & - & 277,956 & 16,090 \\ 
        \hline
    \end{tabular}
\end{table*}

We excluded any manuscripts retrieved that were either review articles or marked as drafts not to be cited. As a continuation and following the search process using the queries in \cref{table:inclusion_exclusion_article}, all the records were merged, duplicates were removed, and a unique ID was assigned to each record. As we reused the word ``cancer" in every search query, some overlapping results were returned as well. 
Based on the inclusion and exclusion criteria of the literature used for the systematic review, and based on the outcome, we used only selected research papers that were more relevant, recent and highly cited. Since the systematic review was conducted a few months back, depending on the contents, addition or deletion to/from the above databases might happen. Thus, the same queries may return a different result. 
The pre-processing task involved, tokenisation, part-of-speech~(PoS) tagging, stemming, removal of stop words, dependency parsing, word sense disambiguation, linking words: 

\begin{enumerate}[noitemsep]
    \item \textbf{Tokenization} involves parsing the text into atomic terms, words, and symbols. 
    \item \textbf{Stemming} is about reducing tokens into their stem, base or root form. 
    \item \textbf{PoS tagging} is used to identify terms representing verbs, nouns, adjectives, etc. 
    \item \textbf{Dependency parsing} about extracting a grammatical tree structure for a sentence where the leaf nodes indicate individual words that together form phrases. 
    \item \textbf{Word sense disambiguation} to identify the meaning in which a word is used. 
    \item \textbf{Linking words} linking terms with a lexicon of senses based on a domain-specific ontology. 
\end{enumerate}
 
Resulting texts are then represented in CoNLL format\footnote{\url{http://universaldependencies.org/docs/format.html}}. 

\subsubsection{Named entity recognition}
The bio-entity extraction component involves two tasks: i) NER that recognizes the named entities in PubMed abstracts based on BioBERT~\cite{BioBERT} and SciBERT~\cite{SciBERT} models, ii) EL to link extracted named entities, and iii) RE. NER is about recognizing domain-specific proper nouns in a biomedical corpus. For example, for the sentence \texttt{The main proteins regulating cell-cycle are BAD and PNCa, which have downstream effects}, proteins, cell-cycle, BAD, and PNCa are 4 proper nouns. \texttt{TP53 and FAS are the top two POTSF genes in terms of the number of associated cancer types, which are associated with 34 and 15 cancer types}, is another example, where a NER model is able to identify that TP53, FAS, POTSF genes, and cancer are the named entities. Recent NER approaches extract such named entities based on learning frameworks that leverage lexical features such as POS tags, dependency parse trees, etc~\cite{hogan2020knowledge}. 

\begin{table*}[h]
    \centering
    \caption{Hyperparameter combination for BERT variants~\cite{karim_phd_thesis_2022}}
    \label{table:bert_params}
    \vspace{-2mm}
    \begin{tabular}{l|l|l|l}
        \hline
        \textbf{Hyperparameter}  & \textbf{BioBERT-cased} &
        \textbf{SciBERT-cased} & \textbf{BERT-cased}\\
        \hline
        Learning rate & 3e-5 & 5e-5 & 2e-5 \\
        Epochs & 6 & 6 & 5 \\
        Max sequence length & 128 & 128 & 128 \\
        Dropout & 0.3 & 0.3 & 0.4 \\
        Batch size & 16 & 16 & 16 \\
        \hline
    \end{tabular}
\end{table*}
    
 While fine-tuning BioBERT and SciBERT, WordPiece tokenization is used in which any new words is represented by frequent sub-words~\cite{wu2016google}. 
In WordPiece embedding, a word is divided into masked tokens. For example, \texttt{syndromes} is represented as \texttt{s \#\#yn \#\#dr \#\#om \#\#es}. 
The compatibility of BioBERT with BERT helps allows representing any new words and fine-tuned for the biomedical domain using the original WordPiece vocabulary of BERT~\cite{BioBERT}. Eventually, it helps  mitigates the out-of-vocabulary issue.
The transformers in all the BERT variants connect the encoders and decoders through self-attention for greater parallelization and reduced training time~\cite{hasan2020knowledge,kim2019neural}. 

\begin{figure*}[h]
	\centering
	\includegraphics[width=0.8\textwidth]{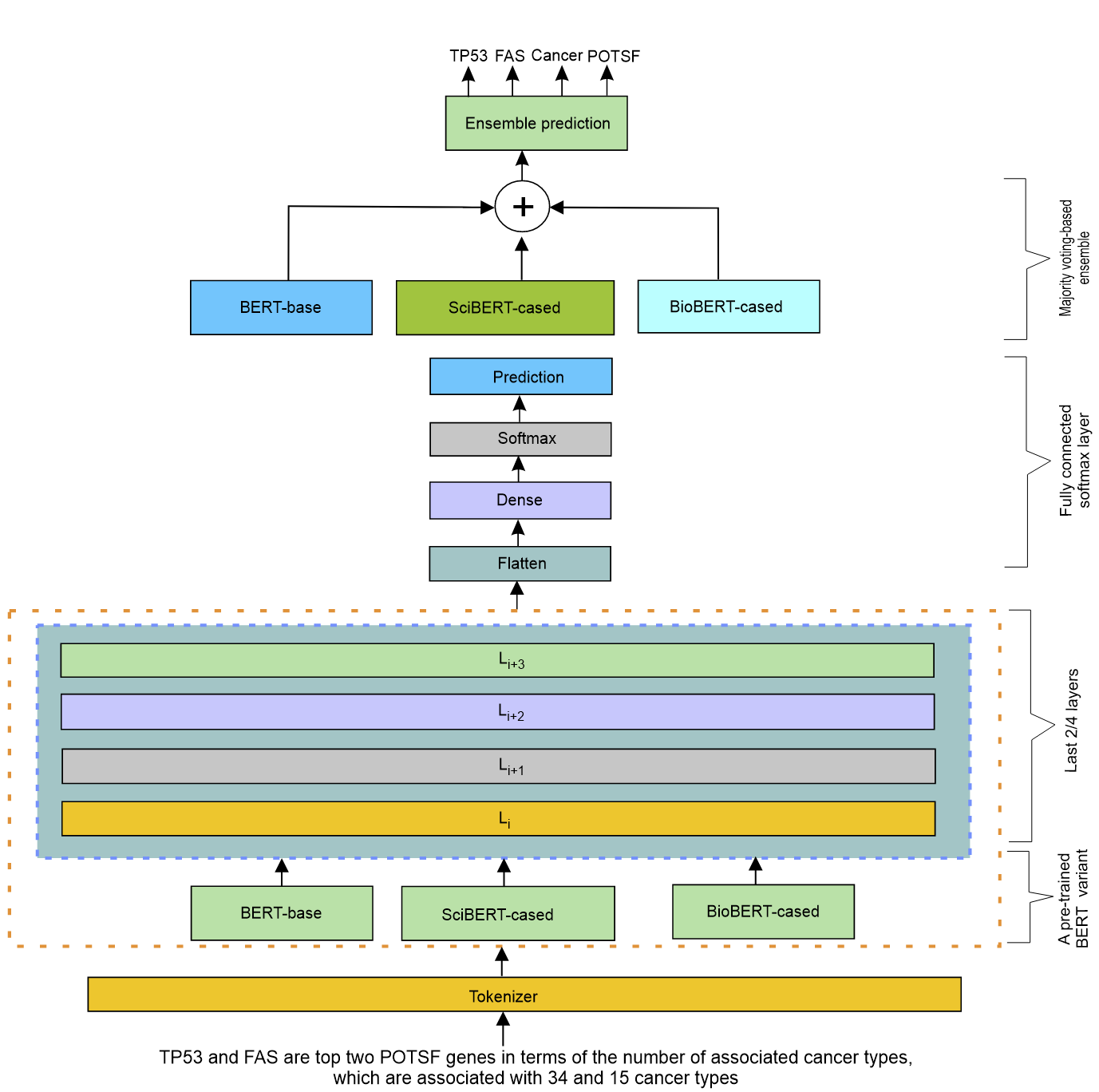}
    \caption[Workflow of NER task from scientific articles]{A schematic representation of NER task. Each of 3 different BERT variants are fine-tuned by adding a fully-connected softmax layer on top for predicting the learned named entities~\cite{karim_phd_thesis_2022}}
	\label{fig:ner_bert_pipeline}
\end{figure*}

BioBERT and SciBERT are initialized with a case-sensitive version of BERT~(i.e., BERT cased). Then SciBERT and BioBERT's weights are pre-trained on PubMed abstracts. Each pre-trained model are further fine-tuned to perform the NER task based on the PubMed Central full-text articles collected based on the inclusion and the exclusion criteria in \cref{table:inclusion_exclusion_article}. Inspired by the NER model proposed by Kim et a.~\cite{kim2019neural}, our SciBERT- and BioBERT-based NER models are trained to predict the following 7 tags~\cite{kim2019neural}: 

\begin{enumerate}[noitemsep]
    \item \textbf{[IOB tags]} - Inside, Outside, and Begin.
    \item \textbf{[X]} - sub-token of WordPiece.
    \item \textbf{[CLS]} - leading token of a sequence for classification. 
    \item \textbf{[SEP]} - a sentence delimiter.
    \item \textbf{[PAD]} - padding of each word in a sentence.
\end{enumerate}

BERT was originally pre-trained on English Wikipedia, news, and book corpus. Thus, it requires domain-specific finetuning for biomedical domain texts that contain a considerable number of domain-specific proper nouns and terms. Recently, BioBERT and SciBERT are recently finetuned for solving different biomedical tasks, as shown in \cref{fig:ner_bert_pipeline}. Both variants for BERT are found to be very effective at recognizing known biomedical entities~\cite{kim2019neural}. The NER models are fine-tuned as follows~\cite{kim2019neural}:

\begin{align}
    \vspace{-6mm}
    p\left(T_{i}\right)=\operatorname{softmax}\left(T_{i} W^{T}+b\right)_{k},
\end{align}

 where $k$ represents the indexes of seven tags $\{\mathrm{B}, \mathrm{I}, \mathrm{O}, \mathrm{X},[\mathrm{CLS}],[\mathrm{SEP}], \mathrm{PAD}\}$; where $p$ is the probability distribution of assigning each $k$ to token $i$, and $T_{i} \in R^{H}$ is the final hidden representation. The final hidden representation is calculated by BioBERT and SciBERT for each token $i$, $H$ is the hidden size of $T_{i}, W \in R^{K \times H}$ is a weight matrix between $k$ and $T_{i}$; where $K$ represents the number of tags, and $b$ is a $\mathrm{K}$-dimensional bias vector on each $k$. The classification loss $L$ is calculated as follows~\cite{kim2019neural}:

\begin{align}
    \vspace{-6mm}
    L(\Theta)=-\frac{1}{N} \sum_{i=1}^{N} \log \left(p\left(y_{i} \mid T_{i}; \Theta\right)\right. ,
    \label{eq:bert_class_loss}
\end{align}

 where $\Theta$ represents model parameters and $N$ is the sequence length. SciBERT and BioBERT extract different entity types, where an entity or two with frequently-occurring token interaction is marked with more than one entity type span. Then, based on probability distribution, we choose the correct entity when they were tagged with more than two types w.r.t probability-based decision rules~\cite{kim2019neural}.

\subsubsection{Entity linking and relation extraction}   
Assuming these entities already exist in the KG, we link the given mentions to these nodes. The EL task also involves named entity disambiguation, e.g., there could be multiple ways to mention the same entity~(e.g., \texttt{TP53} and \texttt{Li-Fraumeni syndrome} are anonymously used). In such a case, the information available under both mentions across different nodes are splitted, under the assumptions that various aliases and multilingual labels are already captured during the NER task. Further, multi-type normalization is performed to assign unique IDs to extracted bio entities. Finally, RE, which is the task of classifying relations of named
entities in a biomedical corpus, is performed. 

We consider both binary and n-ary relation types, in a closed setting between gene/protein and disease relation types. We utilize the sentence classifier of BERT-cased variant. For the RE, BERT uses a [CLS] token for the classification of relations. Similar to BioBERT-based RE, sentence classification is performed using a single output layer based on a [CLS] token representation from BERT. The target named entities in a sentence is anonymized using pre-defined @GENE\$ and @DISEASE\$. For example, the sentence: \emph{TP53 is responsible for breast cancer. TP53 has POTSF functionality, which is mentioned in several PubMed articles} having four target entities is represented as \emph{@GENE\$ is responsible for a disease called @DISEASE\$. @GENE\$ has POTSF functionality, which is mentioned in several PubMed articles}. 

\subsubsection{Integrating extracted triples into KG}
Possible sub-classes of \emph{Significance} is defined as HIGH, MEDIUM, and LOW based on the annotations provided in the TumorPortal\footnote{\url{www.tumorportal.org/tumor_types?ttype=PanCan}}. Further, each entity belonging to the class biomarker is annotated with the properties available in the OGG ontology. Besides, the number of scientific articles is included as an evidence associated with certain genes. 
The \emph{Disease} class consists of a subclass named \emph{DiseaseOfCellularProliferation} under which there is another sub-class called \emph{Cancer}. We included those 33 cancer types of interests in Cancer class, which are labelled with their well-known abbreviations, e.g., BRCA for breast cancer. 
For the entities in the class disease, object properties from the DOID ontology are also inherited and included as annotations. 

\begin{figure*}[h]
	\centering
	\includegraphics[width=0.8\textwidth]{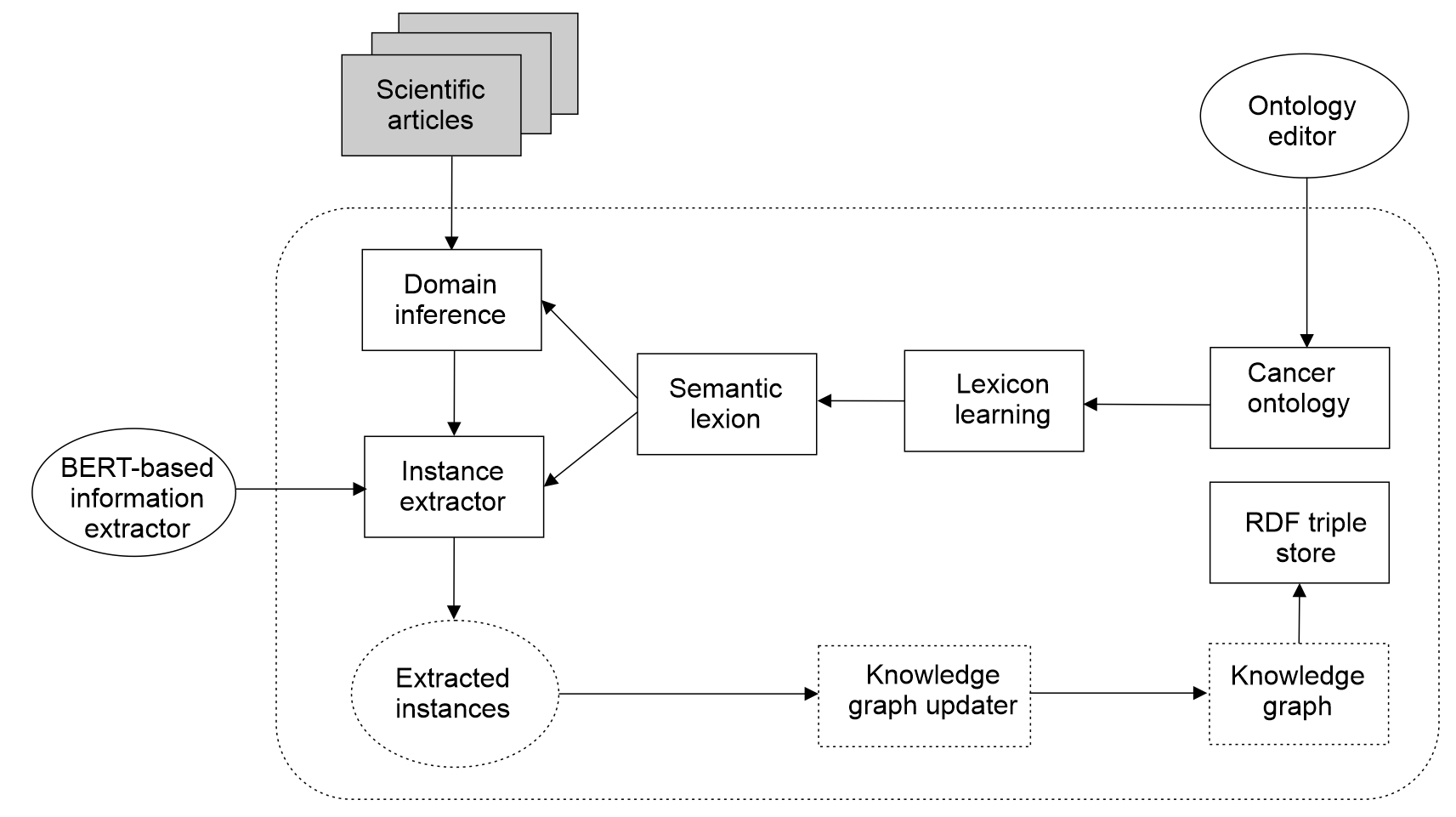}
    \label{fig:kg_enrichment}
	\caption{Knowledge graph enrichment with cancer specific biomarkers~\cite{karim_phd_thesis_2022}}
\end{figure*}

The \emph{Feature} class connects the disease and responsible genes. Besides, the significance degree of a particular gene is included w.r.t a disease. Besides, biomarker and evidence type information are also included, where PubMed, MeSH, and CancerIndex are considered as the source of the evidence. These features are used to indicate the relation between entities from different classes, which in turn annotations allow the rules generation. Since the gene biomarkers can be categorized as \emph{Oncogene}, \emph{ProteinCoding}, and \emph{POTSF}, we considered these as the subclasses of \emph{BiomarkerType}. 

In the RE step, links between extracted lexical terms from the source text and the concepts from the ontology is defined. Then, using a BioBERT and SciBERT-based NER module, the context of the terms will be analyzed to determine the appropriate disambiguation, before assigning the term to the correct concept. Finally, attribute-value pairs are identified, which involves the identification of a subject, mapping it to a semantic class, and using the predicate and object as the attribute name and value, respectively. 


We incorporate a semantic lexicon to integrate new facts into the KG. The lexicon extractor is a BioBERT and SciBERT-based instance extractor, which: i) uses the rules from the ONO ontology to enrich the semantic lexicon with new lexicon symbols and ii) extracts the instance information and create RDF triples in the form of $(u,e,v)=(\mathit{subject},\mathit{predicate},\mathit{object})$, where each triple forms a connected component of a sentence for the KG. For example, for the sample text in \cref{fig:ner_bert_pipeline}, \texttt{TP53 is responsible for a disease called Breast Cancer. TP53 has POTSF functionality, which is mentioned in numerous PubMed articles}, \emph{TP53}, \emph{disease}, \emph{Breast Cancer}, \emph{POTSF}, and \emph{PubMed} are the named entities. This yields the following relation triples: \texttt{(TP53, causes, Breast Cancer), (TP53, hasType, POTSF), (Breast Cancer, isA, Disease), (POTSF, hasEvidence, PubMed)}. 

\subsubsection{Generating explanations for NER}
Similar to literature~\cite{karim2020deephateexplainer}, local explanations for NER task is provided in a post-hoc fashion. Important features in a text sample are identified and highlighted for individual predictions. Relevance score~(RS) as a measure of importance is computed with SA and relevance conservation LRP~\cite{arras2017explaining}. Similar to LRP-based explanation, 
first an input text is processed and the prediction is generated to identify named entities. Then, the output value for each entity predicted is back-propagated layer-to-layer onto the token level, where token relevances are visualized as a heatmap. 
For the input vector of a an instance $x$, the relevance score $R_{d}$ is computed for each input dimension $d$. This is analogous to quantify the relevance of $x_{d}$ w.r.t to target class $c$, where $R_{d}$ is generated by computing the squared partial derivatives as follows~\cite{arras2017explaining}:

\begin{align}
    R_{d}=\left(\frac{\partial f_{c}}{\partial x_{d}}(\boldsymbol{x})\right)^{2},
\end{align}

 where $f_{c}$ is a prediction score function for class $c$. Total relevances is then computed by summing relevances of all input space dimensions $d$~\cite{arras2017explaining}:

\begin{align}
    \left\|\nabla_{\boldsymbol{x}} f_{c}(\boldsymbol{x})\right\|_{2}^{2}.
\end{align}

LRP is based on the layer-wise relevance conservation principle, which redistributes the quantity $f_{c}(\boldsymbol{x})$ from the output layer to the input layer. The relevance for the output layer neuron is set to $f_{c}(\boldsymbol{x})$ w.r.t to the target class $c$, by ignoring irrelevant output layer neurons. The layer-wise relevance score for each intermediate lower-layer neuron is computed based on weighted connections. Suppose $z_{j}$ and $z_{i}$ are an upper-layer and a lower-layer neuron. Assuming the value of $z_{j}$ is already computed in the forward pass as $\sum_{i} z_{i} \cdot w_{ij}+b_{j}$, where $w_{ij}$ and $b_{j}$ are the weight and bias, the relevance score $R_{i}$ for a lower-layer neuron $z_{i}$ is then computed by distributing the relevences onto lower-layer. 

The relevance propagation $R_{i \leftarrow j}$ from upper-layer neurons $z_{j}$ to lower-layer neurons $z_{i}$ is computed as a fraction of the relevance $R_{j}$. All the incoming relevance for each lower-layer neuron is then summed up as~\cite{arras2017explaining}:

\begin{align}
    R_{i \leftarrow j}=\frac{z_{i} \cdot w_{i j}+\frac{\epsilon \cdot \operatorname{sign}\left(z_{j}\right)+\delta \cdot b_{j}}{N}}{z_{j}+\epsilon \cdot \operatorname{sign}\left(z_{j}\right)} \cdot R_{j}
\end{align}

where $N$ is total number of lower-layer neurons connected to $z_{j}$, $\epsilon$ is a stabilizer,  $\operatorname{sign}\left(z_{j}\right)=$ $\left(1_{z_{j} \geq 0}-1_{z_{j}<0}\right)$ is the sign of $z_{j}$, and $\delta$ is a constant multiplicative factor set to 1, to conserve the total relevance of all neurons in the same layer. Finally, $R_{i}$ is computed as $R_{i}= \sum_{j} R_{i \leftarrow j}$~\cite{arras2017explaining}.

\subsection{Quality assessment of knowledge graph}
Once the KG is constructed and enriched with additional facts from external sources, a crucial step is to assess the fitness of purpose. The proliferation of massive KGs poses a question on the quality of the embedded knowledge~(i.e., entities and relations), and whether these facts precisely depict the intended real-world concepts and their relationships. Further, regardless of data sources, the initial KG will usually be incomplete, will contain duplicates, anomalies or errors, contradictory or even incorrect statements – especially when the resultant is based on multiple sources~\cite{hogan2020knowledge}. Therefore, quality assessment is of utmost significance to ensure the quality of the inferred knowledge. 

 \begin{table*}[h]
    \centering
    \caption{Quality assessment metrics~(availability, completeness, conciseness, interlinking, performance, relevancy dimensions) and their interpretation(source: based on literature~\cite{sejdiu2019scalable})}
    \label{table:sansa_qa}
    \vspace{-2mm}
    \begin{tabular}{p{4.45cm}|p{13cm}} 
        \hline
        \rowcolor{Gray}
         \textbf{Metric} & \textbf{Description} \\\hline
            Schema completeness  & Measures the ratio of the number of classes and relations of the gold standard existing in this linked data and the number of classes and relations in the gold standard. \\\hline
            Interlinking completeness & Measures the interlinking completeness. Since any resource of a dataset can be interlinked with another resource of a foreign dataset this metric makes a statement about the ratio of interlinked resources to resources that could potentially be interlinked. \\\hline%
            Property completeness & Measures the property completeness by checking the missing object values for the given predicate and given class of subjects. A user specifies the RDF class and the RDF predicate, then it checks for each pair whether instances of the given RDF class contain the specified RDF predicate. \\\hline
            Numeric range checker & Checks if the incorrect numeric range for the given predicate and given class of subjects. A user should specify the RDF class, the RDF property for which he would like to verify if the values are in the specified range determined by the user. The range is specified by the user by indicating the lower and the upper bound of the value.\\\hline%
            Extensional conciseness & Checks for redundant resources in the assessed dataset, and thus measures the number of unique instances found in the dataset. \\\hline 
            External SameAs links & Checks the sameAs externals links \\\hline 
            Datatypes compatibility  & Checks if the value of a typed literal is valid with regards to the given xsd datatype.  \\\hline
            Dereferenceable URIs & Calculates the number of valid redirects of URIs. It computes the ratio between the number of all valid redirects (subject + predicates + objects) a.k.a dereferenced URIS and the total number of URIs on the dataset. \\\hline
            Dereferenceable back links  & Measures the extent to which a resource includes all triples from the dataset that have the resource's URI as the object. The ratio computed is the number of objects that are "back-links" (are part of the resource's URI) and the total number of objects. \\\hline
            Dereferenceable forward links & Measures the extent to which a resource includes all triples from the dataset that have the resource's URI as the subject. The ratio computed is the number of subjects that are "forward-links" (are part of the resource's URI) and the total number of subjects.  \\\hline
            Coverage detail & Measures the the coverage (i.e. number of entities described in a dataset) and level of detail (i.e. number of properties) in a dataset to ensure that the data retrieved is appropriate for the task at hand. \\\hline
            Coverage scope & calculate the coverage of a dataset referring to the covered scope. This covered scope is expressed as the number of 'instances' statements are made about.	\\\hline%
            Labeled resources  & Assess the labeled resources \\\hline 
    \end{tabular}
\end{table*}

In graph theory, one of the most established approaches to finding the most important nodes is the centrality indicators. Centrality measures such as closeness-, betweenness-, and eigenvector centrality are used to determine the most influential node in a social network. However, these measures do not take into account the node attributes~\cite{zhang2018link}. Zhang et al.~\cite{zhang2018link} showed that a two or three neighbourhood subgraph around a targeted link is sufficient to predict its attribute values. 
Classification and ranking metrics such as Hits@N and mean reciprocal rank, accuracy, precision, recall, and F-score~\cite{Morris2019} have been currently incorporated to evaluate the KG construction and completion in terms of the factuality of the embedded entities and their relations~\cite{karim2019drug}.

However, since such comprehensive quality assessment is subject to other downstream tasks based on KG embeddings, we consider a lightweight measure is taken to evaluate the integrated ontology in order to detect possible inconsistencies using \texttt{OntOlogy Pitfall Scanner!}~(OOPS)~\cite{poveda2014oops}. OOPS helped us: i) the domain or range of a relationship is defined as the intersection of two or more classes. This warning could avoid reasoning problems in case those classes could not share instances, ii) no naming convention is used in the identifiers of the ontology elements. The maintainability, accessibility and clarity of the ontology could be improved, and iii) cycles between two classes in the hierarchy is included in the ontology. As for the KG, we assess its quality w.r.t availability, completeness, conciseness, interlinking, performance, relevancy dimensions. For this, we use SANSA linked data quality assessment metrics~\cite{sejdiu2019scalable} described in \cref{table:sansa_qa}.

\section{Experiment results}\label{chapter_8:results}
In this section, we report some experiment results. 

\subsection{Experiment setup}
For pre-training of \textit{BERT}-based NER models, the maximum input length is set to $256$. During the fine-tuning for 20 epochs, the training set is shuffled for each epoch, gradient clipping is applied, the initial learning rate is set to $2 e^{-5}$, and the Adam optimizer is used to optimize the loss~(i.e., \cref{eq:bert_class_loss}) with the scheduled learning rate. \Cref{table:bert_params} shows the hyperparameter combination for pretraining and finetuning that are empirically produced with random search and 5-fold cross-validation tests. 
We provide some query benchmarking for selected questions. We generate the DLx rules using different reasoners such as Pellet, ELK, and HermiT. Prot{\'e}g{\'e} 5.5.0 is used for the reasoning, w.r.t DLx querying. 

However, ELK does not support object all values from axioms, while HermiT gave some inconsistencies~(giving a different number of rules). Therefore, we report the rules generated with Pellet reasoner only. The inferred rules based on the reasoning are expressed in DLx format. Finally, generated rules are interpreted, followed by decision reasoning with rules. To evaluate the effectiveness of domain-specific potentials of transformer-based approach for information extraction, a comparative analysis with the well-established CRF and Bi-LSTM architectures are provided. 

\subsection{Analysis of information extraction}
NER involves identifying both entity boundaries and entity types. With \emph{exact-match evaluation}, a named entity is considered correctly recognized by the NER model only if both boundaries and type match the provided ground truth. To validate the knowledge extraction from scientific articles using BioBERT and SciBERT, the performance of the entity extraction is evaluated in terms of entity-level precision and recall. While precision measures the ability of a NER system to present only correct entities, whereas recall measures the ability of the NER model to recognize all entities in our corpus. Since both high precision and high recall is desirable, an F1 score was not shown. Besides, results for the state-of-the-art BERT and the BiLSTM-CRF models are also provided as two different baselines. In \cref{table:ner_results}, we report the precision, recall, and F1 scores with the best scores in bold and the second-best scores are underlined. 

\begin{table*}[h!]
    \centering
    \caption[Performance of named entity recognition and normalization models]{Performance of knowledge extraction models from scientific articles}
    \label{table:ner_results}
    \vspace{-2mm}
    \begin{tabular}{l|l|l|l|l} 
    \hline
        & \multicolumn{2}{c|}{Entity extraction} & \multicolumn{2}{c}{Multi-type normalization} \\ 
        \hline
        \textbf{Model} & \textbf{Precision} & \textbf{Recall} & \textbf{Precision} & \textbf{Recall} \\ 
        \hline
        BiLSTM-CRF  & 83.13\% & 82.19\% & 84.57\% & 83.29\% \\ 
        \cline{1-3}\cline{4-5}
        BERT-cased & 87.25\% & 86.65\% & 88.34\% & 87.31\% \\ 
        \hline
        SciBERT-cased & \underline{89.35\%} &  \underline{88.55\%} &  \underline{90.12\%} &  \underline{89.37\%} \\
        \hline
        BioBERT-cased &\textbf{ 91.36\%} & \textbf{90.75\%} & \textbf{91.32}\% & \textbf{91.43\%} \\
        \hline
    \end{tabular}
\end{table*}

\begin{figure*}
	\centering
	\includegraphics[width=\textwidth]{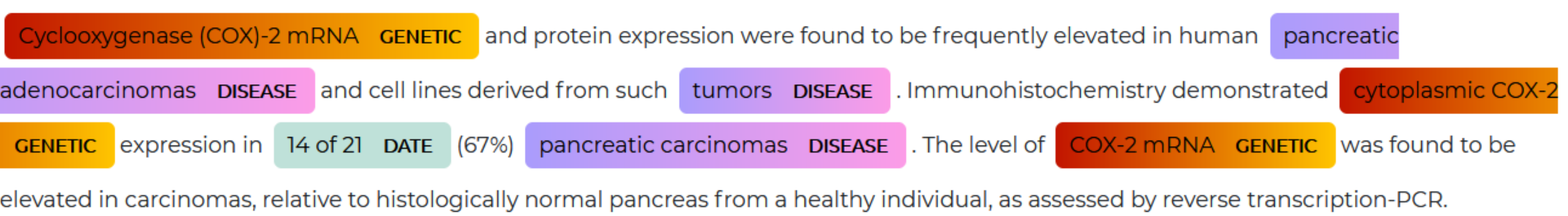}
	\caption{Example of explainable NER, where the model classifies and highlights relevant biomedical entities~\cite{karim2022explainable}}
    \label{fig:bio_ner}
\end{figure*}

BioBERT obtained the highest F1 score in recognizing Genes/Proteins and diseases, which is about 2\% better than SciBERT. BioBERT significantly outperformed BiLSTM-CRF model by $7.85 \%$ w.r.t F1-score. Being fine-tuned on domain-specific articles, SciBERT also outperformed the BiLSTM-CRF by $5.39 \%$ in terms of the F1-score, on average. BERT, which is pre-trained on general domain corpus was highly effective. On average, BioBERT and SciBERT models outperformed BERT by 3.25\% in terms of the F1-score. A sample visual explanation is depicted in \cref{fig:bio_ner}, which highlighted identified genes/proteins and disease-related entities. 

\subsection{Query benchmarks}\label{query_benchmarks}
We foresee the benefit that diagnosis decisions can be validated based on explicitly prior knowledge from the KG. In other words, the KG can be used to provide querying facilities for the end-user in an interactive explanation scenario. Some questions in human language~(NLQ) and DLx query~(DLQ) formats are provided 
in \cref{table:nlq_dlq}. 
For example, answer to NLQ: ``List of all biomarkers which are classified POTSF and responsible for breast cancer answered" against the DLQ ``Biomarker \textcolor{red}{and} causes \textcolor{blue}{some} BRCA \textcolor{red}{and} isA \textcolor{green}{only} POTSF". 
Besides, we outline some SPARQL queries~(ref. \cref{fig:sparql_query_1}, \ref{fig:sparql_query_2}, \ref{fig:sparql_query_3}, \ref{fig:sparql_query_4}, and \ref{fig:sparql_query_5}) that can be used to retrieve domain specific knowledge about biomarkers and associated evidences. Questions in human language~(NLQ) and DLx query~(DLQ) formats are listed in \cref{table:nlq_dlq_2}. 

\begin{table*}
    \caption{Example queries in natural language and DLx formats~\cite{karim2022explainable}}
    \label{table:nlq_dlq}
    \vspace{-4mm}
    \begin{center}
        \scriptsize
        \begin{tabular}{p{7.7cm}|p{10.5cm}}
            \hline
            \rowcolor{Gray}
            \textbf{NLQ} & \textbf{Description logic axioms}\\\hline  
                \begin{flushleft} Which biomarkers have PubMed evidence associated? \end{flushleft} & \begin{flushleft}hasEvidence \textcolor{blue}{some} PubMed \end{flushleft} \\\hline
                
                \begin{flushleft}List of Oncogene, POTSF, and ProteinCoding biomarkers? \end{flushleft} & \begin{flushleft}Biomarker \textcolor{green}{and} isA only Oncogene \textcolor{green}{and} isA only POTSF \textcolor{green}{and} isA only ProteinCoding \end{flushleft}\\\hline
                
                \begin{flushleft}List of Oncogenes responsible for Medulloblastoma with evidence? \end{flushleft} & \begin{flushleft}Biomarker \textcolor{green}{and} causes \textcolor{blue}{some} MED \textcolor{green}{and} isA only Oncogene \textcolor{green}{and} (hasEvidence \textcolor{blue}{some} PubMed \textcolor{red}{or} MeSH) \end{flushleft}\\\hline
                
                \begin{flushleft}List of Oncogene, POTSF, and ProteinCoding genes for breast cancer? \end{flushleft} & \begin{flushleft}Biomarker \textcolor{green}{and} causes \textcolor{blue}{some} BRCA \textcolor{green}{and} isA only Oncogene \textcolor{green}{and} (isA only POTSF \textcolor{red}{or} isA only ProteinCoding) \end{flushleft}\\\hline   
                
                \begin{flushleft}List of Oncogene biomarkers responsible for breast cancer? \end{flushleft} & \begin{flushleft}Biomarker \textcolor{green}{and} causes \textcolor{blue}{some} BRCA \textcolor{green}{and} isA only Oncogene \end{flushleft}\\\hline
                
                \begin{flushleft}List of POTSF biomarkers responsible for breast cancer? \end{flushleft} & \begin{flushleft}Biomarker \textcolor{green}{and} causes \textcolor{blue}{some} BRCA \textcolor{green}{and} isA only POTSF \end{flushleft}\\\hline
                
                \begin{flushleft}List of biomarkers with at least 5 articles as PubMed evidence?\end{flushleft}  & \begin{flushleft}Biomarker \textcolor{green}{and} haveEvidence \textcolor{blue}{some} PubMed \textcolor{green}{and} haveCitations min 5 \end{flushleft}\\\hline
                
                \begin{flushleft}List of biomarkers with maximum 100 articles associated? \end{flushleft} & \begin{flushleft}Biomarker \textcolor{green}{and} haveCitations max 100 \end{flushleft}\\\hline
                
                \begin{flushleft}Which cancer types caused by biomarker ERBB2? \end{flushleft} & \begin{flushleft}Cancer \textcolor{green}{and} causes \textcolor{blue}{some} ERBB2 \end{flushleft}\\\hline
                
                \begin{flushleft}Which cancer types caused by biomarker TP53?  \end{flushleft}& \begin{flushleft}Cancer \textcolor{green}{and} causes \textcolor{blue}{some} TP53 \end{flushleft}\\\hline
                
                \begin{flushleft}Which biomarkers have significance degree High?  \end{flushleft}& \begin{flushleft}Biomarker \textcolor{green}{and} haveSignificance \textcolor{blue}{some} High \end{flushleft}\\\hline
                
                \begin{flushleft}Which biomarkers have significance degree High for BLCA cancer? \end{flushleft} & \begin{flushleft}Biomarker \textcolor{green}{and} causes \textcolor{blue}{some} BLCA \textcolor{green}{and} haveSignificance \textcolor{blue}{some} High \end{flushleft}\\\hline
        \end{tabular}
    \end{center}
\end{table*}

\begin{figure*}[h]
\scriptsize{
    \begin{lstlisting}[frame=bt,numbers=none,language=SPARQL,linewidth=\linewidth,morekeywords={FILTER,AS,GROUP,BY,VALUES,PREFIX}]
    PREFIX ono: <http://www.example.com/ontologies/ono/ono.owl#> 
    PREFIX doid: <http://purl.obolibrary.org/obo/doid#> 
    PREFIX owl: <http://www.w3.org/2002/07/owl#> 

    SELECT DISTINCT ?labelBiomarker WHERE {
                  ?biomarker a owl:Class .
                  ?biomarker rdfs:label ?labelBiomarker .
                  ?disease rdfs:label ?labelDisease .    
                  ?typeEvidence rdfs:label ?evidenceLabel   
                  VALUES ?s {owl:intersectionOf}   ([ rdf:type owl:Restriction ;
                                                   owl:onProperty ?evidence ;
                                                   owl:someValuesFrom ?typeEvidence
                                                    ]
                                                    [ rdf:type owl:Restriction ;
                                                    owl:onProperty ?significance ;
                                                    owl:someValuesFrom ?degree
                                                    ]
                                                    [ rdf:type owl:Restriction ;
                                                    owl:onProperty ?causes ;
                                                    owl:someValuesFrom ?disease
                                                    ]
                                                    [ rdf:type owl:Restriction ;
                                                    owl:onProperty ?hasCitations ;
                                                    owl:cardinality ?number
                                                    ])
            FILTER (regex(?labelDisease, "^BRCA") && regex(?degree, "^High") 
                    && ?number >=100 && regex(?evidenceLabel, "^PubMed"))
                } 
           GROUP BY ?labelBiomarker
    \end{lstlisting}}
    \caption[Query for finding list of biomarkers responsible for breast cancer]{SPARQL query for finding list of biomarkers responsible for breast cancer~(BRCA) that have high significance based on at least 100 citations from PubMed~\cite{karim2022explainable}} 
    \label{fig:sparql_query_1}
\end{figure*}


\begin{figure*}[h]
\scriptsize{
    \begin{lstlisting}[frame=bt,numbers=none,language=SPARQL,linewidth=\linewidth,morekeywords={FILTER,AS,GROUP,BY,VALUES,PREFIX}]
    PREFIX ono: <http://www.example.com/ontologies/ono/ono.owl#> 
    PREFIX doid: <http://purl.obolibrary.org/obo/doid#> 
    PREFIX owl: <http://www.w3.org/2002/07/owl#> 

    SELECT DISTINCT ?labelBiomarker WHERE {
                  ?biomarker a owl:Class .
                  ?biomarker rdfs:label ?labelBiomarker .
                  ?disease rdfs:label ?labelDisease .    
                  VALUES ?s {owl:intersectionOf} (
                                        [ rdf:type owl:Restriction ;
                                       owl:onProperty ?causes ;
                                        owl:someValuesFrom ?disease
                                        ])
            FILTER (regex(?labelDisease, "^BLCA") && regex(?labelDisease, "^CARC") 
            && regex(?labelDisease, "^CRC"))
                } 
            GROUP BY ?labelBiomarker
    \end{lstlisting}}
    \caption[Query for finding list of biomarkers responsible for Urinary Bladder Cancer]{SPARQL query for finding list of biomarkers responsible for Urinary Bladder Cancer\\ and Carcinoid and Colorectal Cancer~\cite{karim2022explainable}} 
    \label{fig:sparql_query_2}
\end{figure*}

\begin{figure*}[h]
\scriptsize{
    \begin{lstlisting}[frame=bt,numbers=none,language=SPARQL,linewidth=\linewidth,morekeywords={FILTER,AS,GROUP,BY,VALUES,PREFIX}]
    PREFIX ono: <http://www.example.com/ontologies/ono/ono.owl#> 
    PREFIX doid: <http://purl.obolibrary.org/obo/doid#> 
    PREFIX owl: <http://www.w3.org/2002/07/owl#> 

    SELECT DISTINCT ?labelBiomarker WHERE {
            SELECT DISTINCT ?labelBiomarker WHERE {
                  ?biomarker a owl:Class .
                  ?biomarker rdfs:label ?labelBiomarker .
                  ?disease rdfs:label ?labelDisease .    
                  ?typeEvidence rdfs:label ?evidenceLabel .
                  ?typeGene rdfs:label ?labelGene   
                  VALUES ?s {owl:intersectionOf} ([rdf:type owl:Restriction ;
                                                    owl:onProperty ?evidence ;
                                                    owl:someValuesFrom ?typeEvidence
                                                    ]
                                                    [ rdf:type owl:Restriction ;
                                                      owl:onProperty doid:isA ;
                                                      owl:allValuesFrom ?typeGene] 
                                                        [ rdf:type owl:Restriction ;
                                                        owl:onProperty ?causes ;
                                                        owl:someValuesFrom ?disease
                                                        ])
        FILTER (regex(?labelDisease, "^Cancer") && regex(?evidenceLabel, "^CancerIndex") 
               && regex(?labelGene, "^Oncogene"))
                } 
        GROUP BY ?labelBiomarker
    \end{lstlisting}}
    \caption[Query for finding list of oncogenes responsible for any type of\\ Cancer]{SPARQL query for finding list of oncogenes responsible for\\ any type of Cancer cited in Cancer Index~\cite{karim2022explainable}} 
    \label{fig:sparql_query_3}
\end{figure*}

\begin{figure*}[h]
\scriptsize{
    \begin{lstlisting}[frame=bt,numbers=none,language=SPARQL,linewidth=\linewidth,morekeywords={FILTER,AS,GROUP,BY,VALUES,PREFIX}]
    PREFIX ono: <http://www.example.com/ontologies/ono/ono.owl#> 
    PREFIX doid: <http://purl.obolibrary.org/obo/doid#> 
    PREFIX owl: <http://www.w3.org/2002/07/owl#> 

    SELECT DISTINCT ?labelBiomarker WHERE {
                  ?biomarker a owl:Class .
                  ?biomarker rdfs:label ?labelBiomarker .
                  ?disease rdfs:label ?labelDisease .    
                  VALUES ?s {owl:intersectionOf} ([ rdf:type owl:Restriction ;
                                                    owl:onProperty ?significance ;
                                                    owl:someValuesFrom ?degree
                                                    ]
                                                    [ rdf:type owl:Restriction ;
                                                    owl:onProperty ?causes ;
                                                    owl:someValuesFrom ?disease
                                                    ])
            FILTER (regex(?labelDisease, "^HNSC") && regex(?labelDisease, "^ESO") 
                    && (regex(?degree, "^High") || (regex(?degree, "^Medium")))
                } 
            GROUP BY ?labelBiomarker

    \end{lstlisting}}
    \caption[Query for finding list of responsible for head and neck and esophageal cancer]{SPARQL query for finding list of responsible for head and neck and esophageal cancer that have at least a medium significance degree~\cite{karim2022explainable}} 
    \label{fig:sparql_query_4}
\end{figure*}

\begin{figure*}[h]
\scriptsize{
    \begin{lstlisting}[frame=bt,numbers=none,language=SPARQL,linewidth=\linewidth,morekeywords={FILTER,AS,GROUP,BY,VALUES,PREFIX}]
    PREFIX ono: <http://www.example.com/ontologies/ono/ono.owl#> 
    PREFIX doid: <http://purl.obolibrary.org/obo/doid#> 
    PREFIX owl: <http://www.w3.org/2002/07/owl#> 

    SELECT DISTINCT ?labelBiomarker WHERE {
                  ?biomarker a owl:Class .
                  ?biomarker rdfs:label ?labelBiomarker .
                  ?disease rdfs:label ?labelDisease .    
                  ?typeEvidence rdfs:label ?evidenceLabel  .
                  ?typeGene rdfs:label ?labelGene   
                
                  VALUES ?s {owl:intersectionOf} ([rdf:type owl:Restriction ;
                                                    owl:onProperty ?evidence ;
                                                    owl:someValuesFrom ?typeEvidence
                                                    ]
                            [ rdf:type owl:Restriction ;
                            owl:onProperty doid:isA ;
                            owl:allValuesFrom ?typeGene] 
                                    [ rdf:type owl:Restriction ;
                                    owl:onProperty ?causes ;
                                    owl:someValuesFrom ?disease
                                        ])
            FILTER (regex(?labelDisease, "^BRCA") && regex(?evidenceLabel, "^PubMed") 
                    && regex(?labelGene, "^Oncogene"))
                } 
            GROUP BY ?labelBiomarker
    \end{lstlisting}}
    \caption[Query for finding list of oncogenes responsible Breast Cancer]{SPARQL query for finding list of oncogenes responsible Breast Cancer based on PubMed~\cite{karim2022explainable}} 
    \label{fig:sparql_query_5}
\end{figure*}


\begin{table*}
    \caption{Example queries in natural language, DLx, and logic formats~\cite{karim2022explainable}}
    \label{table:nlq_dlq_2}
    \vspace{-4mm}
    \begin{center}
        \scriptsize
        \begin{tabular}{p{5cm}|p{5cm}|p{8cm}}
            \hline
            \rowcolor{Gray}
            \textbf{NLQ} & \textbf{Description logic axioms} & \textbf{First-order logic}\\\hline  
                \begin{flushleft}Query for finding list of biomarkers responsible for Urinary Bladder Cancer \end{flushleft}& \begin{flushleft}causes \textcolor{blue}{some} BRCA \textcolor{green}{and} haveSignificance \textcolor{blue}{some} High \textcolor{green}{and} haveCitations min 100 \textcolor{green}{and} haveEvidence \textcolor{blue}{some} PubMed \end{flushleft} & \begin{flushleft} Biomarker $\sqcap$ ($\exists$causes.BreastCancer) $\sqcap$ ($\exists$hasSignificance.High) $\sqcap$ ($\geq$ 100 hasCitations) $\sqcap$ ($\exists$hasEidence.PubMed) $\sqcap$ BreastCancer $\sqsubseteq$ Cancer \end{flushleft} \\\hline
                
                \begin{flushleft} Query for finding list of oncogenes responsible for any type of Cancer cited in Cancer Index \end{flushleft} & \begin{flushleft} causes \textcolor{blue}{some} BLCA  \textcolor{green}{and} causes \textcolor{blue}{some} CARC and causes \textcolor{blue}{some} CRC \end{flushleft} & \begin{flushleft} Biomarker $\sqcap$ ($\exists$causes.UrinaryBladderCancer) $\sqcap$ ($\exists$causes.Carcinoid) $\sqcap$ ($\exists$causes.ColorectalCancer)  $\sqcap$  UrinaryBladderCancer $\sqsubseteq$ Cancer $\sqcap$ Carcinoid $\sqsubseteq$ Cancer  $\sqcap$  ColorectalCancer $\sqsubseteq$ Cancer \end{flushleft} \\\hline
                
                \begin{flushleft}Query for finding list of responsible for head and neck and esophageal cancer that have medium significance degree \end{flushleft} & \begin{flushleft}causes \textcolor{blue}{some} Cancer \textcolor{green}{and} haveEvidence \textcolor{blue}{some} CancerIndex \textcolor{green}{and} isA only Oncogene \end{flushleft}& \begin{flushleft} Biomarker $\sqcap$ ($\exists$causes.Cancer) $\sqcap$ ($\exists$hasEvidence.CancerIndex) $\sqcap$ ($\exists$isA.Oncogene) \end{flushleft} \\\hline
                
                \begin{flushleft}Query for finding list of oncogenes responsible Breast Cancer based on PubMed \end{flushleft} & \begin{flushleft} causes \textcolor{blue}{some} HNSC \textcolor{green}{and} causes \textcolor{blue}{some} ESO \textcolor{green}{and} (hasSignificance \textcolor{blue}{some} Medium or High) \end{flushleft} & \begin{flushleft} Biomarker $\sqcap$ ($\exists$causes.HeadAndNeckCancer) $\sqcap$ ($\exists$causes.EsophagealCancer) $\sqcap$ (($\exists$hasSignificance.High) $\sqcup$ ($\exists$hasSignificance.Medium)) $\sqcap$ HeadAndNeckCancer $\sqsubseteq$ Cancer $\sqcap$ EsophagealCancer $\sqsubseteq$ Cancer \end{flushleft} \\ \hline   
                
                \begin{flushleft}Query for finding list of oncogenes responsible Breast Cancer based on PubMed \end{flushleft} & \begin{flushleft}causes \textcolor{blue}{some} BRCA \textcolor{green}{and} haveEvidence \textcolor{blue}{some} PubMed \textcolor{green}{and} isA only Oncogene \end{flushleft} & \begin{flushleft} Biomarker $\sqcap$ ($\exists$causes.BreastCancer) $\sqcap$ ($\exists$hasEvidence.PubMed) $\sqcap$  ($\exists$isA.Oncogene) $\sqcap$ BreastCancer $\sqsubseteq$ Cancer \end{flushleft}\\
                \hline
            \end{tabular}
    \end{center}
\end{table*}

\subsection{Using the knowledge graph}
Our KG can be viewed as the discrete symbolic representations of knowledge in the form of triples. Therefore, reasoning over this knowledge graph would naturally leverage the symbolic technique and would allowing question answering and reasoning. For instance, given the antecedents: \emph{``All oncogenes are responsible for cancer"; ``TP53 is an oncogene"}, a semantic reasoner would be able to deduce the following fact: \emph{``TP53 is responsible for cancer"}. It would also be possible to perform question answering over the KG. For instance, a user would be able to find tyhe answer for a natural language question such as \emph{``which POTSF biomarker is highly responsible for breast carcinoma and has PubMed evidence?"}, the reasoning engine will follow a logical reasoning path to reason about the concept \emph{`unknown'}, as shown in .

\begin{figure*}[h]
	\centering
		\includegraphics[width=\textwidth]{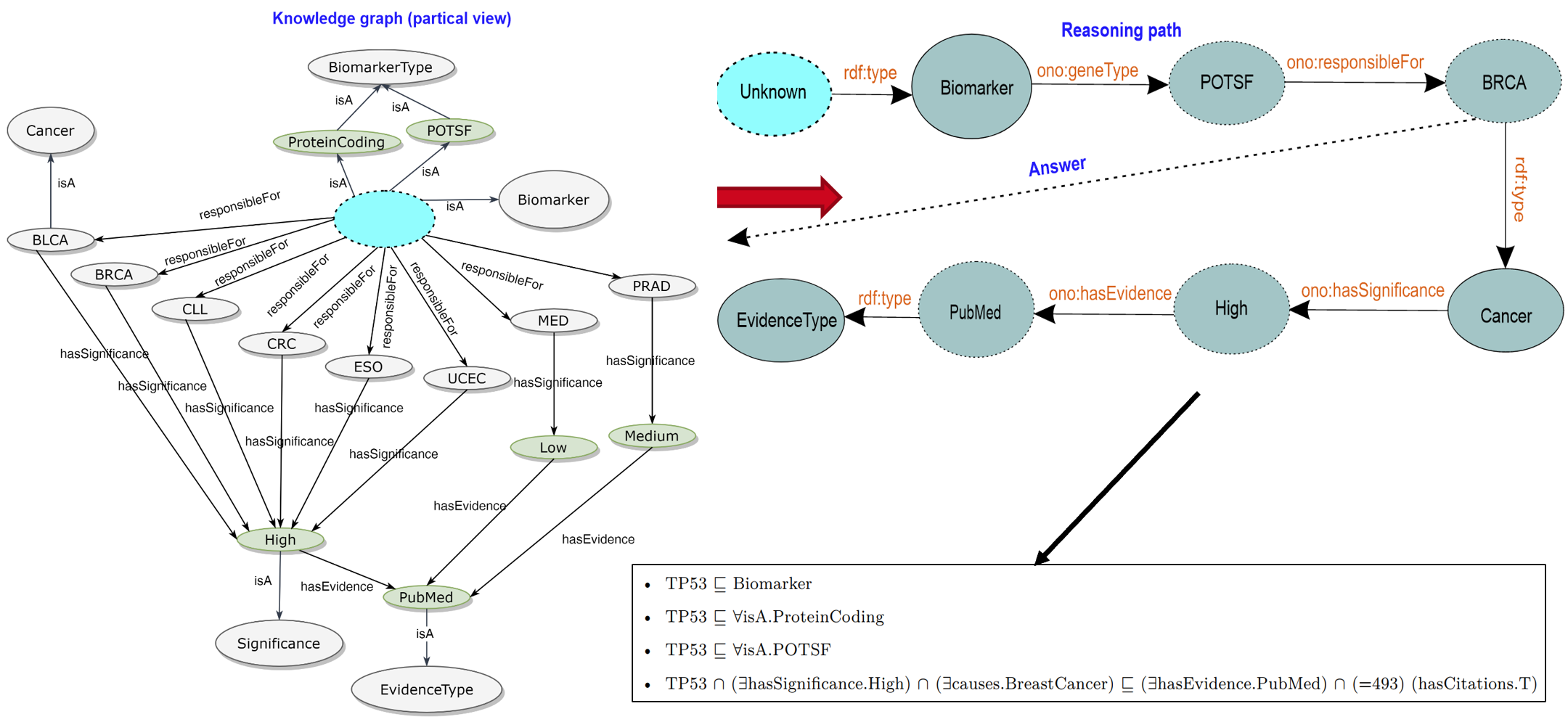}
        \label{fig:ner_lrp}
	\caption{Example of question answering to infer knowledge about named bio-entities~\cite{karim2022explainable}}
\end{figure*}

\begin{figure*}
	\centering
		\includegraphics[width=0.9\textwidth]{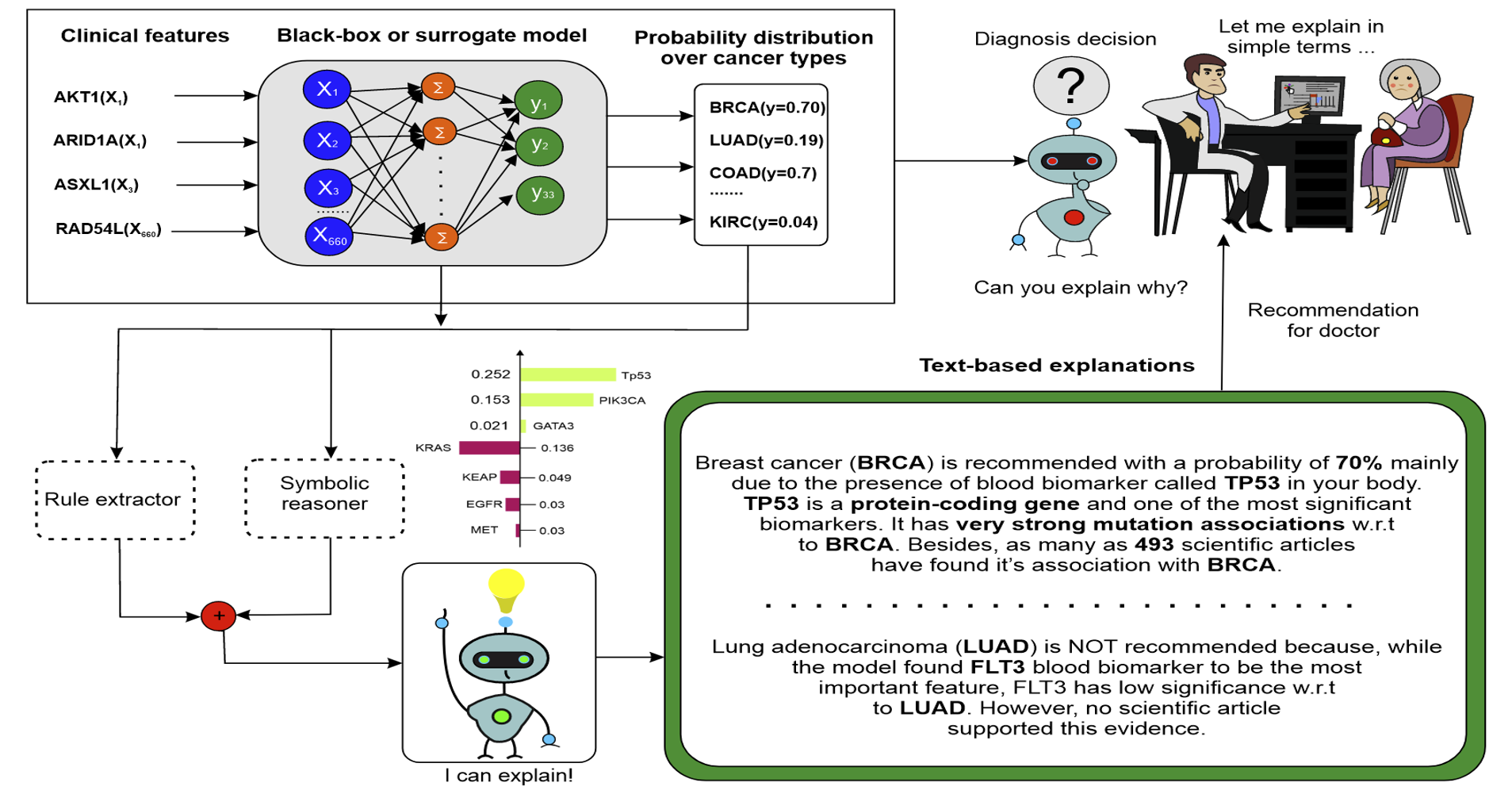}
        \label{fig:decision_reasoning_with_rules}
	\caption{Decision reasoning with rules in textual and natural human language based facts from a knowledge graph~\cite{karim_phd_thesis_2022}}
\end{figure*}

Given the facts about the process and cancer-specific marker genes~(biomarkers), a neuro-symbolic AI system can be developed that would be able to validate the predictions made by the ML model based on the facts presented in the KG. Further, a medical doctor would be bale to explain the diagnosis decision with evidence. A doctor could explain the decision in natural language as well by combining constrastive explanations. \Cref{fig:decision_reasoning_with_rules} shows an example of decision reasoning with rules in textual and natural human language based facts from the KG~\cite{karim2022explainable}. 

\section{Conclusion and Outlook}\label{chapter_8:conclusion}
Our approach based on SW technologies offers a data-efficient process by which models can be trained to reason on symbolic contexts and are able to provide background knowledge for DL models. The semantic layer leverages query processing, instance classification, and ontological reasoning tasks. For these, we developed a domain-specific ontology as the basis of the KB by reusing concepts and knowledge from existing ontologies and KBs. We show how the SR can infer domain knowledge about the top-k biomarkers we already identified. Besides, we validate diagnoses decisions based on domain knowledge from the KB. Generated rules are not only statistically evident but also validated based on the domain-knowledge from several external sources that we embedded in our KG. 

Several potential limitations of our approach leave many improvement possibilities. First, a more comprehensive quality assessment is required to check the correctness of facts in the KG. Further, accuracy that refers to the extent to which entities and relations – encoded by nodes and edges in the graph – correctly represent real-life phenomena~\cite{hogan2020knowledge}, it requires to measure w.r.t syntactic accuracy, semantic accuracy, and timeliness of the KG itself. 
Second, considering their features, symbolic methods are not robust to noise and can not be applied to non-symbolic contexts, where the data is ambiguous~\cite{futia2020integration}. Third, the expressiveness of DLx is not tested against incompleteness and inconsistency. 

Fourth, we could use the fully inferred, deductively closed KG to perform representation learning, e.g., embeddings of nodes and relations. Such KG embeddings on the deductively closed graph would have the advantage that not only asserted axioms will be taken into consideration, but representations can also include inferred knowledge that is not present explicitly in the graph~\cite{alshahrani2017neuro}. 
A complete version of this draft paper about developing a neuro-symbolic AI system will be soon submitted to a journal. 

\bibliographystyle{IEEEtran}
\bibliography{references.bib}

\begin{thebibliography}{10}
\providecommand{\url}[1]{#1}
\csname url@samestyle\endcsname
\providecommand{\newblock}{\relax}
\providecommand{\bibinfo}[2]{#2}
\providecommand{\BIBentrySTDinterwordspacing}{\spaceskip=0pt\relax}
\providecommand{\BIBentryALTinterwordstretchfactor}{4}
\providecommand{\BIBentryALTinterwordspacing}{\spaceskip=\fontdimen2\font plus
\BIBentryALTinterwordstretchfactor\fontdimen3\font minus
  \fontdimen4\font\relax}
\providecommand{\BIBforeignlanguage}[2]{{%
\expandafter\ifx\csname l@#1\endcsname\relax
\typeout{** WARNING: IEEEtran.bst: No hyphenation pattern has been}%
\typeout{** loaded for the language `#1'. Using the pattern for}%
\typeout{** the default language instead.}%
\else
\language=\csname l@#1\endcsname
\fi
#2}}
\providecommand{\BIBdecl}{\relax}
\BIBdecl

\bibitem{ballester2021artificial}
P.~J. Ballester and J.~Carmona, ``Artificial intelligence for the next
  generation of precision oncology,'' pp. 1--3, 2021.

\bibitem{tran2021deep}
K.~A. Tran, O.~Kondrashova, A.~Bradley, E.~D. Williams, J.~V. Pearson, and
  N.~Waddell, ``Deep learning in cancer diagnosis, prognosis and treatment
  selection,'' \emph{Genome Medicine}, vol.~13, no.~1, pp. 1--17, 2021.

\bibitem{phan2016integration}
J.~H. Phan, R.~Hoffman, S.~Kothari, P.-Y. Wu, and M.~D. Wang, ``Integration of
  multi-modal biomedical data to predict cancer grade and patient survival,''
  in \emph{2016 IEEE-EMBS International Conference on Biomedical and Health
  Informatics (BHI)}.\hskip 1em plus 0.5em minus 0.4em\relax IEEE, 2016, pp.
  577--580.

\bibitem{xie2018adaptively}
X.-P. Xie, Y.-F. Xie, Y.-T. Liu, and H.-Q. Wang, ``Adaptively capturing the
  heterogeneity of expression for cancer biomarker identification,'' \emph{BMC
  bioinformatics}, vol.~19, no.~1, p. 401, 2018.

\bibitem{karim2019onconetexplainer}
M.~R. Karim, M.~Cochez, O.~Beyan, S.~Decker, and C.~Lange,
  ``{OncoNetExplainer}: explainable predictions of cancer types based on gene
  expression data,'' in \emph{2019 IEEE 19th International Conference on
  Bioinformatics and Bioengineering (BIBE)}.\hskip 1em plus 0.5em minus
  0.4em\relax IEEE, 2019, pp. 415--422.

\bibitem{karimACCA2019}
M.~R. Karim, O.~Beyan, S.~Decker, and O.~Beyan, ``A snapshot neural ensemble
  method for cancer type prediction based on copy number variations,''
  \emph{Neural Computing and Applications}, vol. 67(1), 2019.

\bibitem{nonakatakeuchi1995}
I.~Nonaka and H.~Takeuchi, \emph{{The Knowledge-Creating Company}}.\hskip 1em
  plus 0.5em minus 0.4em\relax oup, 1995.

\bibitem{xu2020building}
J.~Xu, S.~Kim, M.~Song, M.~Jeong, D.~Kim, J.~Kang, J.~F. Rousseau, X.~Li,
  W.~Xu, V.~I. Torvik \emph{et~al.}, ``Building a pubmed knowledge graph,''
  \emph{Scientific data}, vol.~7, no.~1, pp. 1--15, 2020.

\bibitem{karim_phd_thesis_2022}
\BIBentryALTinterwordspacing
M.~R. Karim, D.~Rebholz-Schuhmann, and S.~Decker, ``Interpreting black-box
  machine learning models for high dimensional datasets,'' Aachen, Germany,
  June 2022. [Online]. Available:
  \url{https://publications.rwth-aachen.de/record/850613}
\BIBentrySTDinterwordspacing

\bibitem{karim2022question}
M.~Karim, H.~Ali, P.~Das, M.~Abdelwaheb, S.~Decker \emph{et~al.}, ``Question
  answering over biological knowledge graph via amazon alexa,'' \emph{arXiv
  preprint arXiv:2210.06040}, 2022.

\bibitem{wilcke2017knowledge}
X.~Wilcke, P.~Bloem, and V.~De~Boer, ``The knowledge graph as the default data
  model for learning on heterogeneous knowledge,'' \emph{Data Science}, vol.~1,
  no. 1-2, pp. 39--57, 2017.

\bibitem{hogan2020knowledge}
A.~Hogan, E.~Blomqvist, M.~Cochez, C.~d'Amato, G.~de~Melo, C.~Gutierrez,
  J.~E.~L. Gayo, S.~Kirrane, S.~Neumaier, A.~Polleres \emph{et~al.},
  ``Knowledge graphs,'' \emph{arXiv preprint arXiv:2003.02320}, 2020.

\bibitem{Liddy.2001}
\BIBentryALTinterwordspacing
E.~D. Liddy, ``Natural language processing,'' \emph{Encyclopedia of Library and
  Information Science, 2nd Ed}, 2001. [Online]. Available:
  \url{http://surface.syr.edu/cgi/viewcontent.cgi?article=1019&context=cnlp}
\BIBentrySTDinterwordspacing

\bibitem{BioBERT}
J.~Lee, W.~Yoon, S.~Kim, D.~Kim, S.~Kim, C.~H. So, and J.~Kang, ``{BioBERT: a
  pre-trained biomedical language representation model for biomedical text
  mining},'' \emph{Bioinformatics}, 2019.

\bibitem{SciBERT}
I.~Beltagy, K.~Lo, and A.~Cohan, ``Scibert: Pretrained language model for
  scientific text,'' in \emph{EMNLP}, 2019.

\bibitem{xue2019fine}
K.~Xue, Y.~Zhou, Z.~Ma, T.~Ruan, H.~Zhang, and P.~He, ``Fine-tuning bert for
  joint entity and relation extraction in chinese medical text,'' in \emph{Int.
  Conf. Bioinformatics and Biomedicine (BIBM)}.\hskip 1em plus 0.5em minus
  0.4em\relax IEEE, 2019, pp. 892--897.

\bibitem{futia2020integration}
G.~Futia and A.~Vetr{\`o}, ``On the integration of knowledge graphs into deep
  learning models for a more comprehensible ai—three challenges for future
  research,'' \emph{Information}, vol.~11, no.~2, p. 122, 2020.

\bibitem{karim2018improving}
M.~R. Karim, A.~Michel, A.~Zappa, P.~Baranov, R.~Sahay, and
  D.~Rebholz-Schuhmann, ``Improving data workflow systems with cloud services
  and use of open data for bioinformatics research,'' \emph{Briefings in
  bioinformatics}, vol.~19, no.~5, pp. 1035--1050, 2018.

\bibitem{wang2015explicit}
P.~Wang, Q.~Wu, C.~Shen, A.~v.~d. Hengel, and A.~Dick, ``Explicit
  knowledge-based reasoning for visual question answering,'' \emph{arXiv
  preprint arXiv:1511.02570}, 2015.

\bibitem{alshahrani2017neuro}
M.~Alshahrani, M.~A. Khan, O.~Maddouri, A.~R. Kinjo, N.~Queralt-Rosinach, and
  R.~Hoehndorf, ``Neuro-symbolic representation learning on biological
  knowledge graphs,'' \emph{Bioinformatics}, vol.~33, no.~17, pp. 2723--2730,
  2017.

\bibitem{hasan2020knowledge}
S.~S. Hasan, D.~Rivera, X.-C. Wu, E.~B. Durbin, J.~B. Christian, and
  G.~Tourassi, ``Knowledge graph-enabled cancer data analytics,'' \emph{IEEE
  journal of biomedical and health informatics}, vol.~24, no.~7, pp.
  1952--1967, 2020.

\bibitem{hu2015semantic}
Q.~Hu, Z.~Huang, and J.~Gu, ``Semantic representation of evidence-based medical
  guidelines and its use cases,'' \emph{Wuhan University Journal of Natural
  Sciences}, vol.~20, no.~5, pp. 397--404, 2015.

\bibitem{POSTF}
L.~Shen, Q.~Shi, and W.~Wang, ``Double agents: genes with both oncogenic and
  tumor-suppressor functions,'' \emph{Oncogenesis}, vol.~7, no.~3, pp. 1--14,
  2018.

\bibitem{slamon1987proto}
D.~J. Slamon, ``Proto-oncogenes and human cancers,'' 1987.

\bibitem{dogan2019fine}
C.~Dogan, A.~Dutra, A.~Gara, A.~Gemma, L.~Shi, M.~Sigamani, and E.~Walters,
  ``Fine-grained named entity recognition using {ELMo} and wikidata,''
  \emph{arXiv preprint arXiv:1904.10503}, 2019.

\bibitem{zheng2017joint}
S.~Zheng, F.~Wang, H.~Bao, Y.~Hao, P.~Zhou, and B.~Xu, ``Joint extraction of
  entities and relations based on a novel tagging scheme,'' \emph{arXiv
  preprint arXiv:1706.05075}, 2017.

\bibitem{zhang2020semi}
M.~Zhang, G.~Geng, and J.~Chen, ``Semi-supervised bidirectional-{LSTM} and
  conditional random fields model for named-entity recognition using embeddings
  from language models representations,'' \emph{Entropy}, vol.~22, no.~2, p.
  252, 2020.

\bibitem{peters2018deep}
M.~E. Peters, M.~Neumann, M.~Iyyer, M.~Gardner, C.~Clark, K.~Lee, and
  L.~Zettlemoyer, ``Deep contextualized word representations,'' \emph{arXiv
  preprint arXiv:1802.05365}, 2018.

\bibitem{sun2020biomedical}
C.~Sun, Z.~Yang, L.~Wang, Y.~Zhang, H.~Lin, and J.~Wang, ``Biomedical named
  entity recognition using bert in the machine reading comprehension
  framework,'' \emph{arXiv preprint arXiv:2009.01560}, 2020.

\bibitem{devlin2018bert}
J.~Devlin, M.-W. Chang, and K.~e.~a. Lee, ``Bert: Pre-training of deep
  bidirectional transformers for language understanding,'' \emph{arXiv preprint
  arXiv:1810.04805}, 2018.

\bibitem{Beltagy2019SciBERT}
I.~Beltagy, K.~Lo, and A.~Cohan, ``Scibert: Pretrained language model for
  scientific text,'' in \emph{EMNLP}, 2019.

\bibitem{chithrananda2020chemberta}
S.~Chithrananda, G.~Grand, and B.~Ramsundar, ``Chemberta: Large-scale
  self-supervised pretraining for molecular property prediction,'' \emph{arXiv
  preprint arXiv:2010.09885}, 2020.

\bibitem{lee2020biobert}
J.~Lee, W.~Yoon, S.~Kim, D.~Kim, S.~Kim, C.~H. So, and J.~Kang, ``Biobert: a
  pre-trained biomedical language representation model for biomedical text
  mining,'' \emph{Bioinformatics}, vol.~36, no.~4, pp. 1234--1240, 2020.

\bibitem{anantharangachar2013ontology}
R.~Anantharangachar, S.~Ramani, and S.~Rajagopalan, ``Ontology guided
  information extraction from unstructured text,'' \emph{arXiv preprint
  arXiv:1302.1335}, 2013.

\bibitem{abu2020domain}
B.~Abu-Salih, ``Domain-specific knowledge graphs: A survey,'' \emph{arXiv
  preprint arXiv:2011.00235}, 2020.

\bibitem{wu2016google}
Y.~Wu, M.~Schuster, Z.~Chen, Q.~V. Le, M.~Norouzi, W.~Macherey, M.~Krikun,
  Y.~Cao, Q.~Gao, K.~Macherey \emph{et~al.}, ``Google's neural machine
  translation system: Bridging the gap between human and machine translation,''
  \emph{arXiv preprint arXiv:1609.08144}, 2016.

\bibitem{kim2019neural}
D.~Kim, J.~Lee, C.~H. So, H.~Jeon, M.~Jeong, Y.~Choi, W.~Yoon, M.~Sung, and
  J.~Kang, ``A neural named entity recognition and multi-type normalization
  tool for biomedical text mining,'' \emph{IEEE Access}, vol.~7, pp.
  73\,729--73\,740, 2019.

\bibitem{karim2022explainable}
M.~Karim, T.~Islam, O.~Beyan, C.~Lange, M.~Cochez, D.~Rebholz-Schuhmann,
  S.~Decker \emph{et~al.}, ``Explainable ai for bioinformatics: Methods, tools,
  and applications,'' \emph{arXiv preprint arXiv:2212.13261}, 2022.

\bibitem{karim2020deephateexplainer}
M.~Karim, S.~K. Dey, B.~R. Chakravarthi \emph{et~al.}, ``Deephateexplainer:
  Explainable hate speech detection in under-resourced bengali language,''
  \emph{arXiv preprint arXiv:2012.14353}, 2020.

\bibitem{arras2017explaining}
L.~Arras, G.~Montavon, K.-R. M{\"u}ller, and W.~Samek, ``Explaining recurrent
  neural network predictions in sentiment analysis,'' \emph{arXiv preprint
  arXiv:1706.07206}, 2017.

\bibitem{sejdiu2019scalable}
G.~Sejdiu, A.~Rula, J.~Lehmann, and H.~Jabeen, ``A scalable framework for
  quality assessment of rdf datasets,'' in \emph{The Semantic Web--ISWC 2019:
  18th International Semantic Web Conference, Auckland, New Zealand, October
  26--30, 2019, Proceedings, Part II 18}.\hskip 1em plus 0.5em minus
  0.4em\relax Springer, 2019, pp. 261--276.

\bibitem{zhang2018link}
M.~Zhang and Y.~Chen, ``Link prediction based on graph neural networks,'' in
  \emph{Advances in Neural Information Processing Systems}, 2018, pp.
  5165--5175.

\bibitem{Morris2019}
\BIBentryALTinterwordspacing
C.~Morris, M.~Ritzert, M.~Fey, W.~Hamilton, J.~E. Lenssen, G.~Rattan, and
  M.~Grohe, ``Weisfeiler and leman go neural: Higher-order graph neural
  networks,'' in \emph{33rd Conf. Artificial Intelligence}, vol.
  4602-4609.\hskip 1em plus 0.5em minus 0.4em\relax {AAAI}, 2019. [Online].
  Available: \url{http://arxiv.org/pdf/1810.02244v2}
\BIBentrySTDinterwordspacing

\bibitem{karim2019drug}
M.~R. Karim, M.~Cochez, and J.~B. e.~a. Jares, ``Drug-drug interaction
  prediction based on knowledge graph embeddings and convolutional-lstm
  network,'' in \emph{Proceedings of the 10th ACM International Conference on
  Bioinformatics, Computational Biology and Health Informatics}.\hskip 1em plus
  0.5em minus 0.4em\relax Niagara Falls, NY, USA: ACM, 2019, pp. 113--123.

\bibitem{poveda2014oops}
M.~Poveda-Villal{\'o}n, A.~G{\'o}mez-P{\'e}rez, and M.~C. Su{\'a}rez-Figueroa,
  ``{OOPS! (OntOlogy Pitfall Scanner!): An On-line Tool for Ontology
  Evaluation},'' \emph{International Journal on Semantic Web and Information
  Systems (IJSWIS)}, vol.~10, no.~2, pp. 7--34, 2014.

\end{thebibliography}

\end{document}